\documentclass[lettersize,journal]{IEEEtran}
\usepackage{subcaption}
\usepackage{orcidlink}
\usepackage{amsmath,amsfonts}
\usepackage{algorithmic}
\usepackage{algorithm}
\usepackage{array}
\usepackage{textcomp}
\usepackage{stfloats}
\usepackage{url}
\usepackage{verbatim}
\usepackage{graphicx}
\usepackage{cite}
\usepackage{multirow}
\usepackage{hyperref}

\begin{document}

\title{Benchmarking Randomized Optimization Algorithms on Binary, Permutation, and Combinatorial Problem Landscapes}

\author{Jethro Odeyemi~\orcidlink{0009-0001-5170-9357}, Wenjun Zhang~\orcidlink{0000-0001-7973-8769},~\IEEEmembership{Senior Member,~IEEE}

\thanks{Jethro Odeyemi is a doctoral candidate at the University of Saskatchewan conducting research at the Advanced Engineering Design Laboratory. (email: jethro.odeyemi@usask.ca)}
\thanks{Wenjun Zhang (Senior Member, IEEE) received the Ph.D. degree from Delft University of Technology, Delft, The Netherlands, in 1994. He is currently a Full Professor with the Department of Mechanical Engineering and the Division of Biomedical Engineering, University of Saskatchewan, Saskatoon, SK, Canada. (email: chris.zhang@usask.ca)}
\thanks{The code and data generated during the course of this study are available on GitHub at \url{https://github.com/jethroodeyemi/comparative-analysis-of-optimization-algorithms}.}
\thanks{This research is supported by the Natural Sciences and Engineering Research Council of Canada (NSERC) Collaborative Research and Training Experience (CREATE) program, grant number: 565429-2022.}}


\maketitle

\begin{abstract}
In this paper, we evaluate the performance of four randomized optimization algorithms—Randomized Hill Climbing (RHC), Simulated Annealing (SA), Genetic Algorithms (GA), and MIMIC (Mutual Information Maximizing Input Clustering)—across three distinct types of problems: binary, permutation, and combinatorial. We systematically compare these algorithms using a set of benchmark fitness functions that highlight the specific challenges and requirements of each problem category. Our study analyzes each algorithm’s effectiveness based on key performance metrics, including solution quality, convergence speed, computational cost, and robustness. Results show that while MIMIC and GA excel in producing high-quality solutions for binary and combinatorial problems, their computational demands vary significantly. RHC and SA, while computationally less expensive, demonstrate limited performance in complex problem landscapes. The findings offer valuable insights into the trade-offs between different optimization strategies and provide practical guidance for selecting the appropriate algorithm based on the type of problems, accuracy requirements, and computational constraints.
\end{abstract}


\section{Introduction}
\IEEEPARstart{O}{ptimization} problems are central to decision-making processes in a wide array of fields such as operations research \cite{Padamwar2019, ji2024pid, ji2020biobjective, ji2020average, liu2012aggregation}, machine learning \cite{Abdulkadirov2023, GAMBELLA2021807, Odeyemi2024}, economics \cite{PONSTEIN1984255}, logistics \cite{GARCIA2015153, doi:10.1080/17517575.2023.2200767}, and engineering design \cite{cao2015unified, cao2018topology, wang2013evacuation, wang2018supplychain}.
 The goal in optimization is to determine the values of decision variables that either maximize or minimize an objective function, often while satisfying a set of constraints. In machine learning, optimization is critical during the model training process, where the objective is typically to minimize a loss function \cite{Bian2024} that measures the error between predicted and true values. The process of tuning model parameters is often accomplished using gradient-based methods such as gradient descent \cite{Baldi1995} and its variants (e.g., stochastic gradient descent \cite{Tian2023}, Adam \cite{Kingma2014AdamAM}). These methods are highly efficient for convex problems, but they face significant challenges in more complex scenarios, particularly when the objective function \cite{Matsuoka2019} is non-convex, riddled with local minima, or spans a high-dimensional space. To overcome these limitations, random optimization algorithms which uses randomness to explore the solution space more thoroughly have emerged as powerful alternatives. Random optimization techniques are particularly effective for solving binary \cite{Becerra-Rozas2023}, permutation \cite{Hu2003}, and combinatorial problems \cite{Peres2021}, where traditional gradient-based methods may not be applicable or efficient. 

This study aims to compare the performance of different optimization algorithms across benchmark functions categorized by solution structure into binary, permutation, and combinatorial problems. The final results highlight trade-offs between computational cost, solution quality (fitness), and convergence speed in selecting appropriate optimization algorithms. We show that Genetic Algorithm (GA) and Mutual Information Maximizing Input Clustering (MIMIC) show strengths in binary problems, while MIMIC excels in permutation problems requiring high solution quality. In contrast, GA performs well in combinatorial problems, offering a balance between accuracy and efficiency. Low-computation-cost algorithms like Randomized Hill climbing (RHC) are useful when resources are limited, albeit with lower solution quality.

The rest of the paper is organized as follows. In Section 2, we discuss related works, reviewing studies that evaluate optimization algorithms across different problem types. Section 3 introduces the randomized optimization algorithms used in this research, along with their specific characteristics. In Section 4, we classify the fitness problems into binary, permutation, and combinatorial categories, explaining the benchmark problems used for evaluation. Section 5 describes the experimental setup, including the environment and tools used for the experiments. Section 6 presents the results and discussion, focusing on the performance of the algorithms across various types of problems. Finally, Section 7 concludes the paper with key findings and future research directions.

\section{Related Works}
Several studies have explored the performance of optimization algorithms across diverse problem types. For example, Derya (2020) evaluated Stochastic Gradient Descent (SGD), Adaptive Gradient Algorithm (AdaGrad), Adaptive Delta (AdaDelta), Adaptive Moment Estimation (Adam), and Root Mean Square Propagation (RMSProp) algorithms in deep learning \cite{Owoeye2023}, focusing on how these algorithms differ in their working principles, strengths, and limitations \cite{Soydaner2020}. There were notable distinctions in their ability to handle noisy gradients, with Adam and RMSProp excelling in adapting learning rates, making them particularly suited for problems involving sparse data and non-convex loss landscapes. In another study, the performance of GA and Simulated Annealing (SA) in maximizing the thermal conductance of harmonic lattices is studied, highlighting the strengths of GA in global exploration, while noting SA’s compute efficiency \cite{KERR201931}. Similarly, a comparative analysis of Ant Colony Optimization (ACO) and Particle Swarm Optimization (PSO) for distance optimization is presented \cite{GUPTA2020245}.

In this study, we analyze optimization challenges within the categories of binary, permutation, and combinatorial problems. In context of permutation problems, a recent investigation explored evolutionary diversity optimization applied to the Traveling Salesperson Problem (TSP) and Quadratic Assignment Problem (QAP) \cite{Do_2022}. The study demonstrated that mutation operators for these permutation problems can ensure convergence towards maximally diverse populations, provided the population size is sufficiently small. Additionally, a Binary version of Equilibrium Optimization (BEO) has been proposed for tackling the 0–1 Knapsack discrete optimization problem \cite{ABDELBASSET2021106946}. Another study introduced the Global Neighborhood Algorithm (GNA), which balances global and local search in optimization \cite{Alazzam2013}.

\section{Randomized Optimization Algorithms}
One of the key challenges in optimization lies in selecting the appropriate algorithm for a given problem type. The effectiveness of an algorithm depends on factors like whether the problem involves continuous or discrete variables, whether the landscape is convex or non-convex, and the complexity of the constraints. Therefore, choosing the right algorithm is crucial to balancing exploration and exploitation of the search space, avoiding local minima, and achieving near-optimal solutions, see also \cite{wang2013predatory}. In this section, we explain the four most widely used randomized optimization algorithms: Randomized Hill Climbing (RHC), Simulated Annealing (SA), Genetic Algorithms (GA), and MIMIC (Mutual Information Maximizing Input Clustering). Each of these algorithms introduces randomness in different ways to enhance exploration and avoid getting trapped in local minima.
\subsection{Randomized Hill Climbing (RHC)}
RHC \cite{liu2016banditbasedrandommutationhillclimbing} is a local search algorithm that iteratively improves a candidate solution by exploring its neighborhood. The algorithm randomly selects a neighbor of the current solution and moves to that neighbor if it offers a better objective function value. If no better neighbors are found, the algorithm stops. RHC introduces randomness by restarting the search from a random point when it reaches a local optimum, which helps explore different regions of the search space. RHC uses the following working principle, summarized below:

\begin{enumerate}
    \item \textbf{Start:} Initialize a random solution $x_0$.
    \item \textbf{Neighbor Selection:} At each iteration, randomly select a neighboring solution $x' \in N(x)$ from the neighborhood of the current solution $x$.
    \item \textbf{Move:} If the objective function $f(x')$ is better than $f(x)$, move to $x'$, as shown in Equation~(1):
    \begin{equation}
    x \leftarrow x'.
    \label{eq:move}
    \end{equation}
    \item \textbf{Restart:} When no improvement is found, restart the algorithm from a new random solution.
\end{enumerate}

At each iteration $t$, the update rule is defined as follows (see Equation~\ref{eq:update_rule}):

\begin{equation}
x_{t+1} = 
\begin{cases} 
x_t & \text{if } f(x_t) \geq f(x') \\ 
x' & \text{if } f(x') > f(x_t)
\end{cases}
\label{eq:update_rule}
\end{equation}

Here, $x' \in N(x_t)$ is a randomly selected neighbor of the current solution $x_t$. RHC is simple but can easily get stuck in local minima without proper restarts \cite{selman2006hill}.

\subsection{Simulated Annealing (SA)}
SA \cite{bertsimas1993simulated} is an extension of hill climbing that introduces a probability of accepting worse solutions to escape local minima. The algorithm is inspired by the annealing process in metallurgy, where materials are slowly cooled to achieve a low-energy configuration. The acceptance of worse solutions is controlled by a temperature parameter $T$, which decreases over time. As the temperature decreases, the algorithm becomes more focused on exploiting local improvements.

\begin{enumerate}
    \item \textbf{Start:} Initialize with a random solution $x_0$ and a high temperature $T_0$.
    \item \textbf{Neighbor Selection:} At each iteration, select a neighboring solution $x' \in N(x)$.
    \item \textbf{Acceptance Criterion:} Move to $x'$ if it improves the objective function ($f(x') > f(x)$). If $f(x') \leq f(x)$, accept $x'$ with probability given by Equation~\ref{eq:accept_prob}:
    \begin{equation}
    P(\text{accept}) = \exp\left(\frac{f(x') - f(x)}{T}\right),
    \label{eq:accept_prob}
    \end{equation}
    where $T$ is the current temperature.
    \item \textbf{Cooling Schedule:} Reduce the temperature according to a cooling schedule, typically as shown in Equation~\ref{eq:cooling_schedule}:
    \begin{equation}
    T(t) = T_0 \times \alpha^t,
    \label{eq:cooling_schedule}
    \end{equation}
    where $\alpha \in (0, 1)$ is the cooling factor and $t$ is the iteration number.
\end{enumerate}

Simulated Annealing allows the search to explore a wide range of solutions at high temperatures and gradually focuses on the best solutions as the temperature decreases.

\subsection{Genetic Algorithms (GA)}
GA \cite{lambora2019genetic} are inspired by the process of natural selection, where a population of solutions evolves over time. GAs maintain a population of candidate solutions, which are recombined and mutated to explore the solution space.

\begin{enumerate}
    \item \textbf{Initialization:} Generate an initial population $P$ of candidate solutions randomly.
    \item \textbf{Selection:} Select parent solutions from the population based on their fitness $f(x)$. Higher-fitness solutions are more likely to be selected.
    \item \textbf{Crossover:} Combine two parent solutions $x_1$ and $x_2$ to create offspring $x_{\text{child}}$ using a crossover operator. For example, in single-point crossover, the offspring is created as shown in Equation~\ref{eq:crossover}:
    \begin{equation}
    x_{\text{child}}[i] =
    \begin{cases} 
    x_1[i] & \text{if } i \leq \text{cross-point} \\ 
    x_2[i] & \text{if } i > \text{cross-point}
    \end{cases}
    \label{eq:crossover}
    \end{equation}
    \item \textbf{Mutation:} Apply mutation to the offspring with some probability to introduce randomness and explore new areas of the solution space.
    \item \textbf{Replacement:} Replace the least-fit members of the population with the new offspring.
\end{enumerate}

The population evolves over time, and the fittest individuals are selected to produce the next generation. The algorithm terminates after a fixed number of generations or when the population converges.

\subsection{MIMIC (Mutual Information Maximizing Input Clustering)}
MIMIC \cite{de1996mimic} is a probabilistic optimization algorithm that constructs a probabilistic model of the solution space and uses this model to generate new candidate solutions. MIMIC builds a dependency tree based on mutual information between variables and samples solutions from this model.

\begin{enumerate}
    \item \textbf{Initialization:} Generate an initial population $P$ of random solutions.
    \item \textbf{Model Building:} Construct a probabilistic model of the population by estimating the mutual information between variables. This model captures dependencies between decision variables.
    \item \textbf{Sampling:} Generate new solutions by sampling from the probabilistic model. These new solutions are used to explore the search space.
    \item \textbf{Selection:} Retain the top $k$ solutions based on their fitness and use these to update the probabilistic model.
\end{enumerate}

MIMIC balances exploration and exploitation by using probabilistic models to sample promising regions of the search space, making it highly efficient for large combinatorial problems.

\textbf{Mutual Information} between two variables $X_i$ and $X_j$ is given by Equation~\ref{eq:mutual_information}:

\begin{equation}
I(X_i; X_j) = \sum_{x_i, x_j} p(x_i, x_j) \log \frac{p(x_i, x_j)}{p(x_i)p(x_j)},
\label{eq:mutual_information}
\end{equation}

where $I(X_i; X_j)$ is the mutual information and $p(x_i, x_j)$ is the joint probability distribution.

\section{Experiments}
\subsection{Experimental Setting}
The experiments were implemented using Python version 3.11.9 and the mlrose \cite{Hayes19} package to apply various search algorithms to randomized optimization problems. All simulations were conducted on a system equipped with an AMD Radeon RX 6800S GPU, 15.6 GB of usable memory, and executed in an interactive notebook environment. 

The primary objective of this study was to analyze the behavior of different algorithms across separate problem groups, focusing on the quantitative aspects of their performance rather than optimizing hyperparameters. Therefore, we used the default parameters provided in the mlrose library as a baseline and tested slight variations around these defaults.

Table \ref{tab:hyperparameters} summarizes the hyperparameters for each algorithm and the variations used during the experiments. For \textbf{SA}, we used the default exponential decay schedule with an exponential constant (\texttt{ExpConst}) of 0.005 and experimented with different values: 0.001, 0.005, and 0.01. For \textbf{GA}, the default population size was set to 200, and we tested additional population sizes of 100 and 300. Similarly, for \textbf{MIMIC}, the population size was set to 200 by default, with variations of 100 and 300 tested.An exception was made for \textbf{RHC}, where the default restarts value was 0. To better understand the effect of restarts, we experimented with values of 0, 5, and 10.

\begin{table}[h!]
    \centering
    \caption{Hyperparameter Settings for SA, RHC, GA and MIMIC Algorithms}
    \begin{tabular}{|p{1cm}|p{3cm}|p{3cm}|}
        \hline
        \textbf{Algorithm} & \textbf{Default Setting} & \textbf{Tested Variations} \\ \hline
        \textbf{SA} & Exponential Decay, ExpConst = 0.005 & ExpConst = 0.001, 0.005, 0.01 \\ \hline
        \textbf{GA} & Population size = 200 & Population size = 100, 200, 300 \\ \hline
        \textbf{MIMIC} & Population size = 200 & Population size = 100, 200, 300 \\ \hline
        \textbf{RHC} & Restarts = 0 & Restarts = 0, 5, 10 \\ \hline
    \end{tabular}
    \label{tab:hyperparameters}
\end{table}

\subsection{Performance Metrics}
During each run, the following performance metrics were monitored:
\begin{itemize}
    \item \textbf{Fitness Score}: The objective value for each problem, representing how well the algorithm optimized the solution \cite{kozeny2015genetic}.
    \item \textbf{Time Complexity}: The time taken for the algorithm to converge to a solution.
    \item \textbf{Convergence Behavior}: The count of function evaluation (feval) required for the algorithm to converge to an optimal or near-optimal solution.
\end{itemize}

To ensure reliable results, we averaged the values of these metrics over five experiment runs. This averaging reduces the impact of random variations and provides a more robust comparison of algorithmic performance across different problem groups.
\section{Results and Discussion}
\subsection{Binary Problems}
\begin{figure*}[htbp]
    \centering
    \begin{subfigure}[b]{0.49\textwidth}
        \centering
        \includegraphics[width=\textwidth]{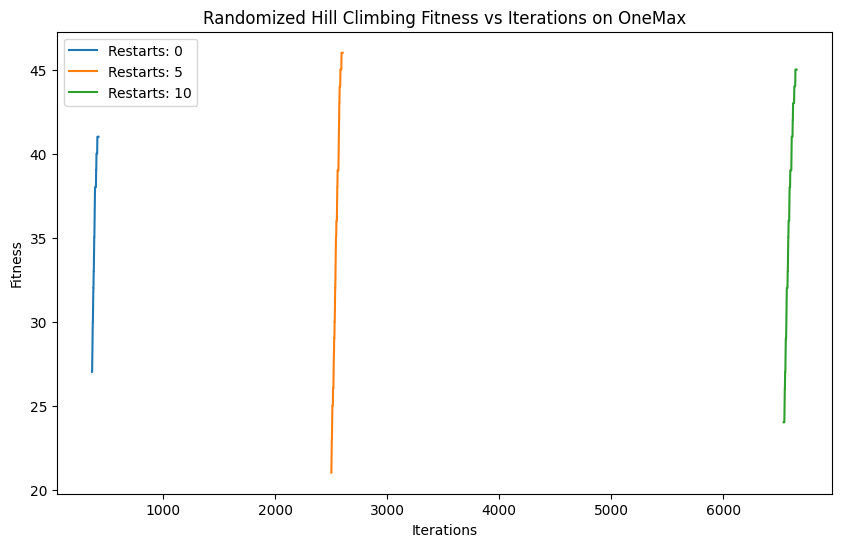}
        \caption{}
        \label{fig:onemaxrhc}
    \end{subfigure}
    \hfill
    \begin{subfigure}[b]{0.49\textwidth}
        \centering
        \includegraphics[width=\textwidth]{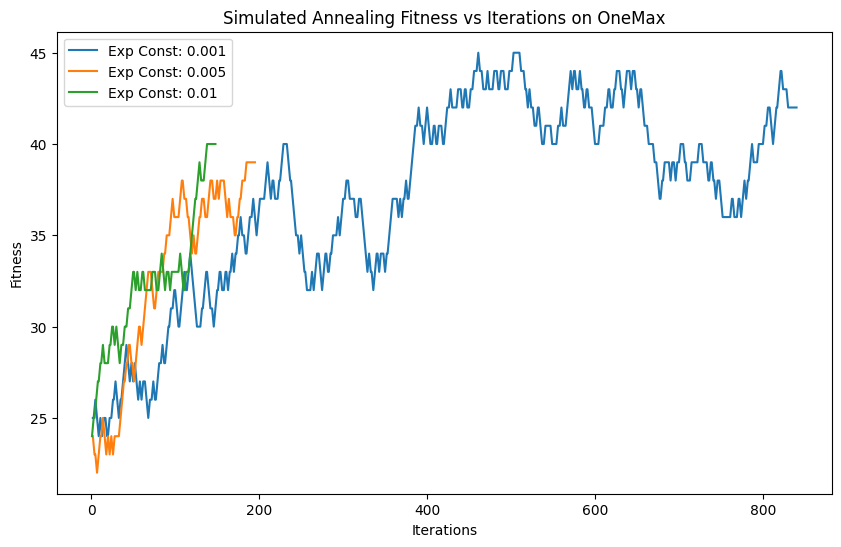}
        \caption{}
        \label{fig:onemaxsa}
    \end{subfigure}
    \vskip\baselineskip
    \begin{subfigure}[b]{0.49\textwidth}
        \centering
        \includegraphics[width=\textwidth]{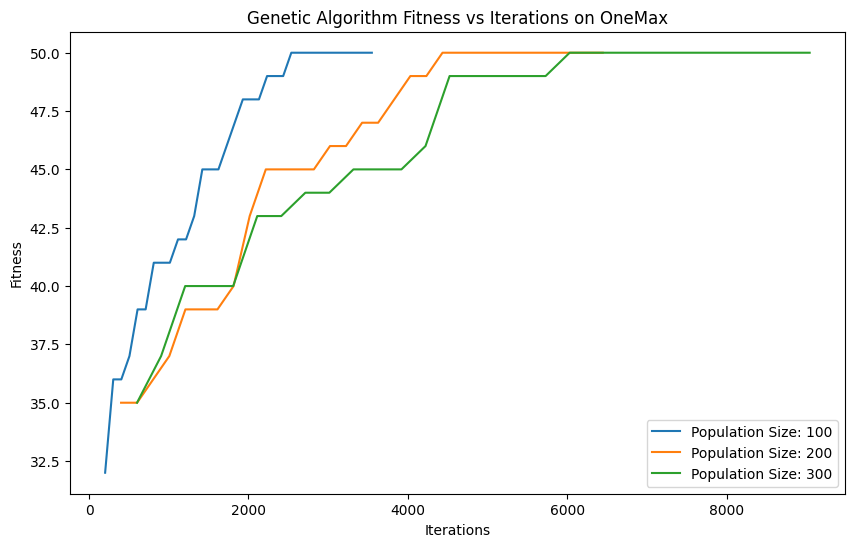}
        \caption{}
        \label{fig:onemaxga}
    \end{subfigure}
    \hfill
    \begin{subfigure}[b]{0.49\textwidth}
        \centering
        \includegraphics[width=\textwidth]{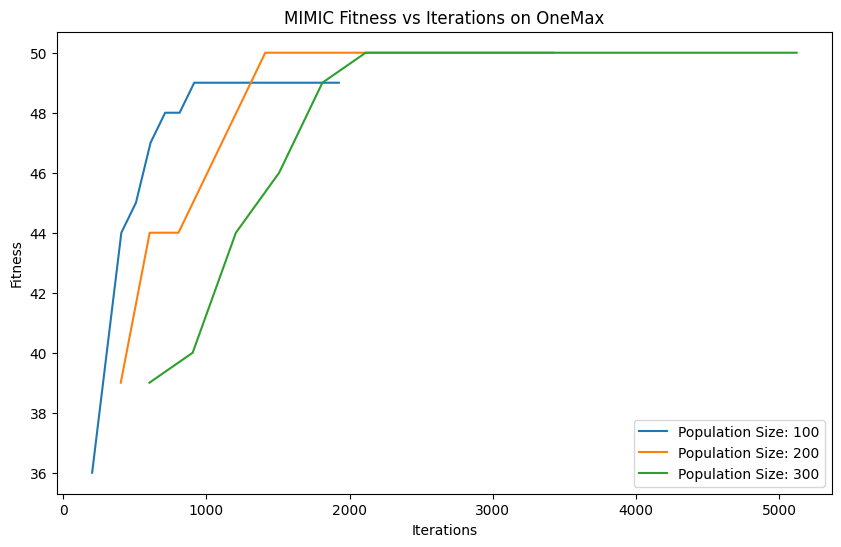}
        \caption{}
        \label{fig:onemaxmimic}
    \end{subfigure}
    \caption{Fitness vs. iterations for the OneMax using different randomized optimization algorithms: (a) RHC, (b) SA, (c) GA, and (d) MIMIC. The plots show the impact of varying key parameters—restarts for RHC, exponential cooling for SA, population sizes for GA, and population sizes for MIMIC—on the fitness progression over iterations.}
    \label{fig:onemaxgrid}
\end{figure*}

For the OneMax problem, where the objective is to maximize the number of 1's in a binary string, the performance of the algorithms reflects their ability to efficiently explore a simple and highly structured search space. Both GA and MIMIC perform exceptionally well, consistently achieving the maximum fitness of 50 across all runs (as seen in Table~\ref{tab:onemax_results}). This is because the OneMax problem has a smooth fitness landscape, where small incremental improvements reliably guide the algorithm toward the global optimum. GA's crossover and mutation operators, enable efficient exploitation of this structure, while MIMIC's probabilistic sampling method effectively concentrates on high-quality solutions early \cite{de1996mimic}, resulting in faster convergence, as illustrated in Figures \ref{fig:onemaxga} and \ref{fig:onemaxmimic}. On the other hand, RHC shows improvement with more restarts. This limitation arises because, without sufficient restarts, RHC lacks the global exploration required to escape local optima in the binary search space (Figure~\ref{fig:onemaxrhc}). Given that the OneMax landscape provides a clear and direct path to the global optimum, SA’s probabilistic moves add unnecessary noise, making it less efficient compared to GA and MIMIC, which can better exploit the straightforward nature of the search space (Figure~\ref{fig:onemaxsa}).
\begin{table}[h!]
    \centering
    \caption{OneMax Problem Results for RHC, SA, GA, and MIMIC with Different Hyperparameters}
    \begin{tabular}{|c|c|c|c|}
        \hline
        \textbf{Algorithm} & \textbf{Param} & \textbf{Average Fitness} & \textbf{Feval} \\ \hline
        
        \textbf{RHC} & Restarts = 0 & 43.4 & 45 \\ \hline
        \textbf{RHC} & Restarts = 5 & 46.8 & 77 \\ \hline
        \textbf{RHC} & Restarts = 10 & 47.6 & 95 \\ \hline
        
        \textbf{SA} & ExpConst = 0.001 & 40.4 & 576 \\ \hline
        \textbf{SA} & ExpConst = 0.005 & 42.0 & 131 \\ \hline
        \textbf{SA} & ExpConst = 0.01 & 41.6 & 104 \\ \hline
        
        \textbf{GA} & PopSize = 100 & 50.0 & 34 \\ \hline
        \textbf{GA} & PopSize = 200 & 50.0 & 31 \\ \hline
        \textbf{GA} & PopSize = 300 & 50.0 & 29 \\ \hline
        
        \textbf{MIMIC} & PopSize = 100 & 49.8 & 18 \\ \hline
        \textbf{MIMIC} & PopSize = 200 & 50.0 & 16 \\ \hline
        \textbf{MIMIC} & PopSize = 300 & 50.0 & 16 \\ \hline
        
    \end{tabular}
    \label{tab:onemax_results}
\end{table}

\begin{figure*}[htbp]
    \centering
    \begin{subfigure}[b]{0.49\textwidth}
        \centering
        \includegraphics[width=\textwidth]{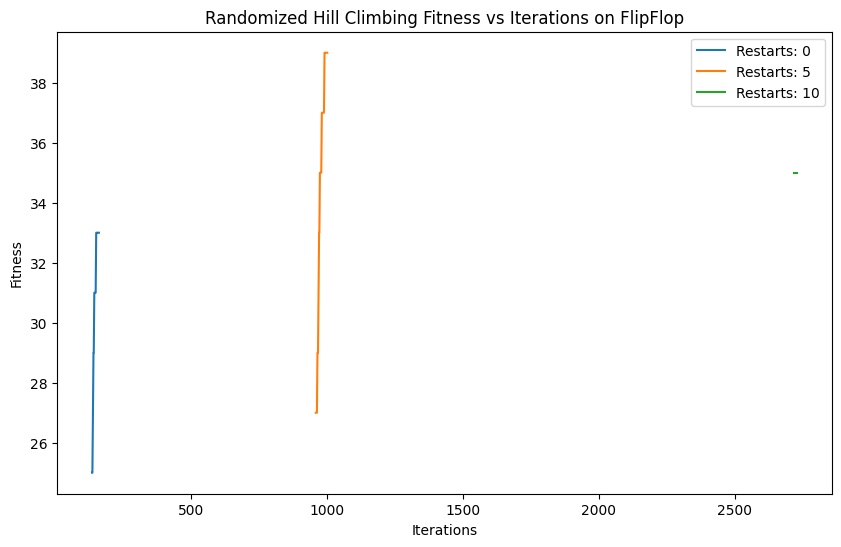}
        \caption{}
        \label{fig:flipfloprhc}
    \end{subfigure}
    \hfill
    \begin{subfigure}[b]{0.49\textwidth}
        \centering
        \includegraphics[width=\textwidth]{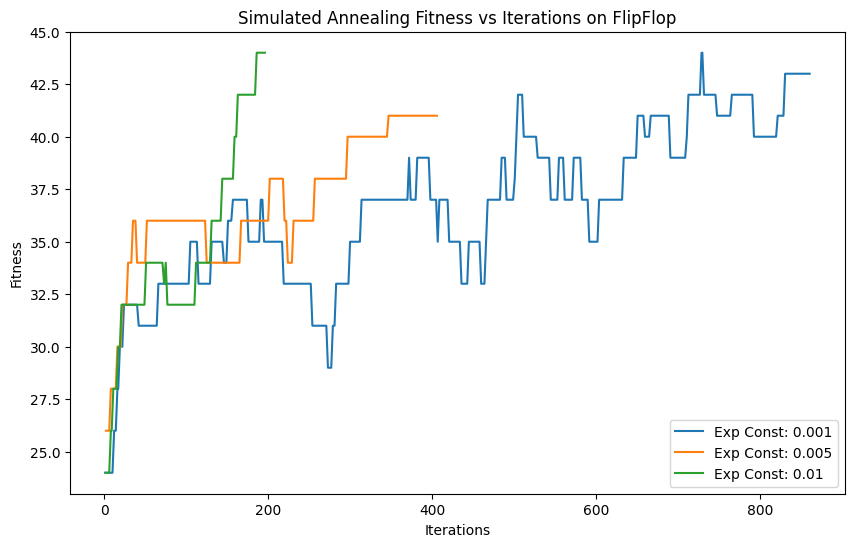}
        \caption{}
        \label{fig:flipflopsa}
    \end{subfigure}
    \vskip\baselineskip
    \begin{subfigure}[b]{0.49\textwidth}
        \centering
        \includegraphics[width=\textwidth]{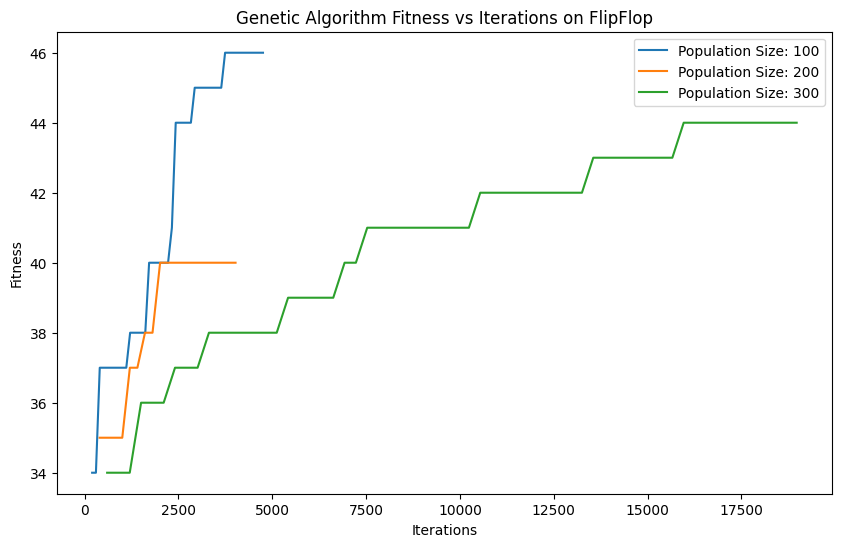}
        \caption{}
        \label{fig:flipflopga}
    \end{subfigure}
    \hfill
    \begin{subfigure}[b]{0.49\textwidth}
        \centering
        \includegraphics[width=\textwidth]{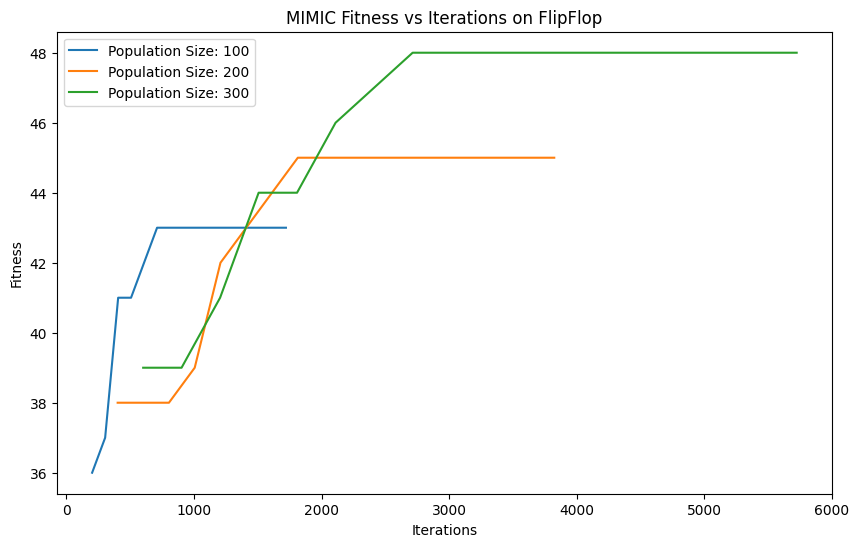}
        \caption{}
        \label{fig:flipflopmimic}
    \end{subfigure}
    \caption{Fitness vs. iterations for the FlipFlop using different randomized optimization algorithms: (a) RHC, (b) SA, (c) GA, and (d) MIMIC. The plots show the impact of varying key parameters—restarts for RHC, exponential cooling for SA, population sizes for GA, and population sizes for MIMIC—on the fitness progression over iterations.}
    \label{fig:flipflopgrid}
\end{figure*}

In the FlipFlop problem, where the objective is to maximize the number of alternating sequences of 1’s and 0’s in a binary string, the performance of the algorithms exhibits both similarities and differences compared to the OneMax problem. GA and MIMIC continue to perform strongly, with GA achieving an reaching a maximum fitness of 43.8 for a population size of 100 and MIMIC closely matching with a fitness of 46.8 at population size of 300 (Table~\ref{tab:flipflop_results}). This is because both GA and MIMIC are well-suited to explore the more rugged landscape of the FlipFlop problem, which includes more local optima than the smooth OneMax problem, as seen in Figures ~\ref{fig:flipflopga} and ~\ref{fig:flipflopmimic}. In contrast, RHC struggles significantly more with FlipFlop, achieving only 33.0 fitness without restarts, and improving marginally to 37.6 with 5 restarts, indicating that RHC's inability to effectively escape local optima is even more pronounced in this problem (Table ~\ref{tab:flipflop_results}, Figure ~\ref{fig:flipfloprhc}). SA, although somewhat improved compared to OneMax, still suffers from inconsistencies in performance due to its probabilistic acceptance of worse solutions, with fitness values fluctuating between 41.2 and 43.0 depending on the cooling schedule (Table ~\ref{tab:flipflop_results}, Figure ~\ref{fig:flipflopsa}). This fluctuation shows that while SA’s flexibility allows it to explore the more complex landscape, the inherent noise in its search process continues to prevent it from reliably reaching optimal solutions, especially when compared to the more structured approaches of GA and MIMIC.
\begin{table}[h!]
    \centering
    \caption{FlipFlop Problem Results for RHC, SA, GA, and MIMIC with Different Hyperparameters}
    \begin{tabular}{|c|c|c|c|}
        \hline
        \textbf{Algorithm} & \textbf{Param} & \textbf{Average Fitness} & \textbf{Feval} \\ \hline
        
        \textbf{RHC} & Restarts = 0 & 33.0 & 22 \\ \hline
        \textbf{RHC} & Restarts = 5 & 37.6 & 37 \\ \hline
        \textbf{RHC} & Restarts = 10 & 36.6 & 10 \\ \hline
        
        \textbf{SA} & ExpConst = 0.001 & 41.8 & 587 \\ \hline
        \textbf{SA} & ExpConst = 0.005 & 41.2 & 277 \\ \hline
        \textbf{SA} & ExpConst = 0.01 & 43.0 & 130 \\ \hline
        
        \textbf{GA} & PopSize = 100 & 43.8 & 46 \\ \hline
        \textbf{GA} & PopSize = 200 & 43.0 & 19 \\ \hline
        \textbf{GA} & PopSize = 300 & 43.0 & 62 \\ \hline
        
        \textbf{MIMIC} & PopSize = 100 & 42.8 & 16 \\ \hline
        \textbf{MIMIC} & PopSize = 200 & 45.6 & 18 \\ \hline
        \textbf{MIMIC} & PopSize = 300 & 46.8 & 18 \\ \hline
        
    \end{tabular}
    \label{tab:flipflop_results}
\end{table}

\begin{figure*}[htbp]
    \centering
    \begin{subfigure}[b]{0.49\textwidth}
        \centering
        \includegraphics[width=\textwidth]{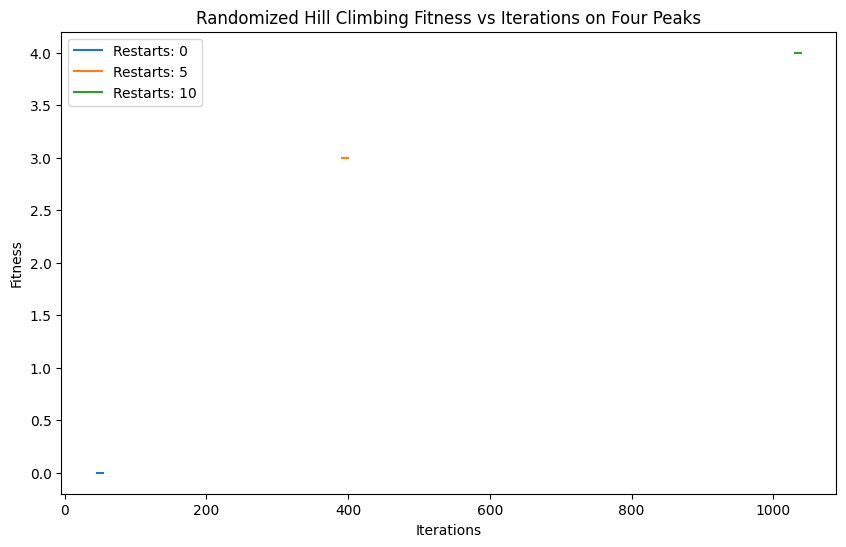}
        \caption{}
        \label{fig:fourpeakrhc}
    \end{subfigure}
    \hfill
    \begin{subfigure}[b]{0.49\textwidth}
        \centering
        \includegraphics[width=\textwidth]{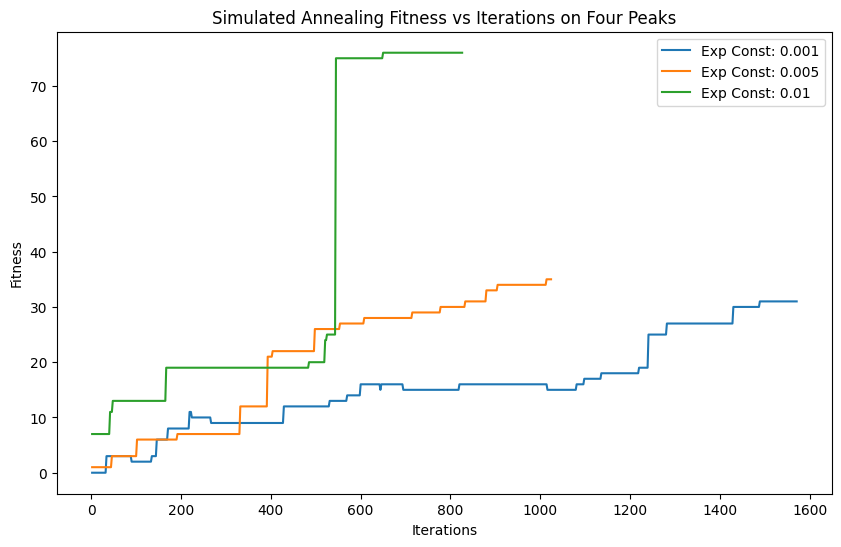}
        \caption{}
        \label{fig:fourpeaksa}
    \end{subfigure}
    \vskip\baselineskip
    \begin{subfigure}[b]{0.49\textwidth}
        \centering
        \includegraphics[width=\textwidth]{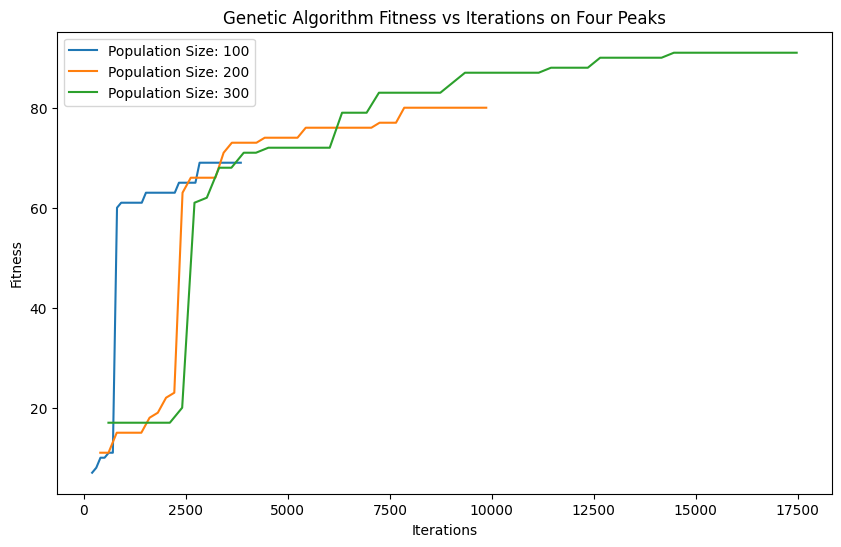}
        \caption{}
        \label{fig:fourpeakga}
    \end{subfigure}
    \hfill
    \begin{subfigure}[b]{0.49\textwidth}
        \centering
        \includegraphics[width=\textwidth]{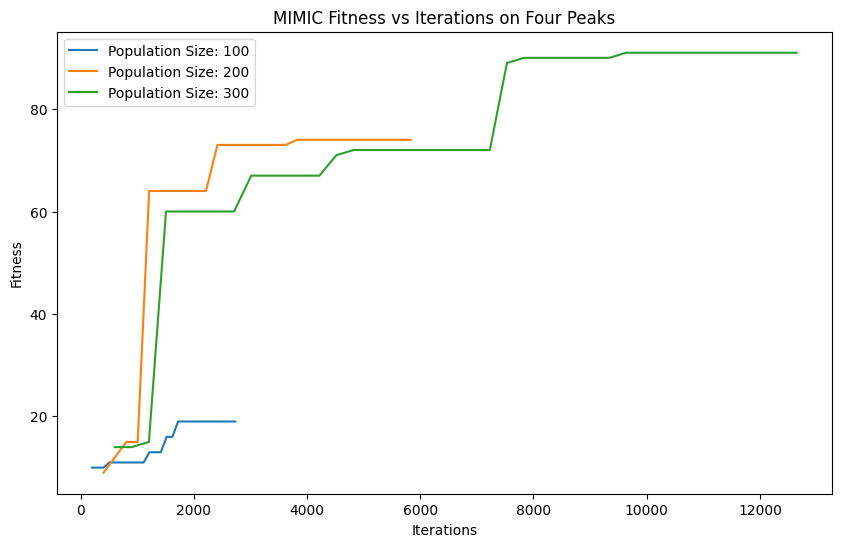}
        \caption{}
        \label{fig:fourpeakmimic}
    \end{subfigure}
    \caption{Fitness vs. iterations for the FourPeaks using different randomized optimization algorithms: (a) RHC, (b) SA, (c) GA, and (d) MIMIC. The plots show the impact of varying key parameters—restarts for RHC, exponential cooling for SA, population sizes for GA, and population sizes for MIMIC—on the fitness progression over iterations.}
    \label{fig:fourpeakgrid}
\end{figure*}

In the FourPeaks problem, where the goal is to balance between finding both long runs of 1's and 0's in a binary string, the performance of the algorithms highlights the challenge of navigating a more complex landscape with multiple local optima. GA and MIMIC again perform well, with GA achieving a high average fitness of 85.8 for a population size of 300 (Table~\ref{tab:fourpeaks_results}), and MIMIC close behind at 68.4 with the same population size. As seen in Figures~\ref{fig:fourpeakga} and~\ref{fig:fourpeakmimic}, both algorithms quickly converge to near-optimal solutions, but GA is notably faster in reaching higher fitness values, demonstrating its advantage in balancing exploration and exploitation for this problem's complex landscape. In contrast, RHC struggles significantly, achieving a maximum fitness of only 6.0 with 10 restarts (Table~\ref{tab:fourpeaks_results}, Figure~\ref{fig:fourpeakrhc}), as it cannot effectively escape local optima in the FourPeaks landscape, where more global exploration is crucial. SA, while showing better results than RHC, has highly variable performance depending on the cooling schedule, with the best average fitness of 39.8 when using an exponential constant of 0.01 (Table~\ref{tab:fourpeaks_results}, Figure~\ref{fig:fourpeaksa}). SA's ability to accept worse solutions helps in this problem, but its stochastic nature introduces instability in performance, making it less reliable than GA or MIMIC.
\begin{table}[h!]
    \centering
    \caption{FourPeaks Problem Results for RHC, SA, GA, and MIMIC with Different Hyperparameters}
    \begin{tabular}{|c|c|c|c|}
        \hline
        \textbf{Algorithm} & \textbf{Param} & \textbf{Average Fitness} & \textbf{Feval} \\ \hline
        
        \textbf{RHC} & Restarts = 0 & 2.2 & 10 \\ \hline
        \textbf{RHC} & Restarts = 5 & 5.8 & 10 \\ \hline
        \textbf{RHC} & Restarts = 10 & 6.0 & 10 \\ \hline
        
        \textbf{SA} & ExpConst = 0.001 & 29.0 & 941 \\ \hline
        \textbf{SA} & ExpConst = 0.005 & 31.4 & 645 \\ \hline
        \textbf{SA} & ExpConst = 0.01 & 39.8 & 538 \\ \hline
        
        \textbf{GA} & PopSize = 100 & 35.4 & 37 \\ \hline
        \textbf{GA} & PopSize = 200 & 66.4 & 48 \\ \hline
        \textbf{GA} & PopSize = 300 & 86.8 & 57 \\ \hline
        
        \textbf{MIMIC} & PopSize = 100 & 34.2 & 26 \\ \hline
        \textbf{MIMIC} & PopSize = 200 & 60.2 & 28 \\ \hline
        \textbf{MIMIC} & PopSize = 300 & 68.4 & 41 \\ \hline
        
    \end{tabular}
    \label{tab:fourpeaks_results}
\end{table}

\begin{figure*}[htbp]
    \centering
    \begin{subfigure}[b]{0.49\textwidth}
        \centering
        \includegraphics[width=\textwidth]{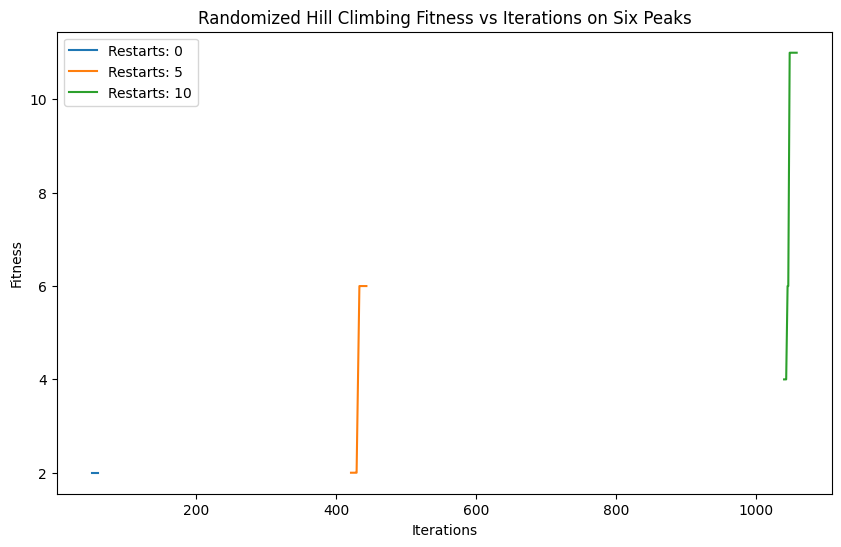}
        \caption{}
        \label{fig:sixpeakrhc}
    \end{subfigure}
    \hfill
    \begin{subfigure}[b]{0.49\textwidth}
        \centering
        \includegraphics[width=\textwidth]{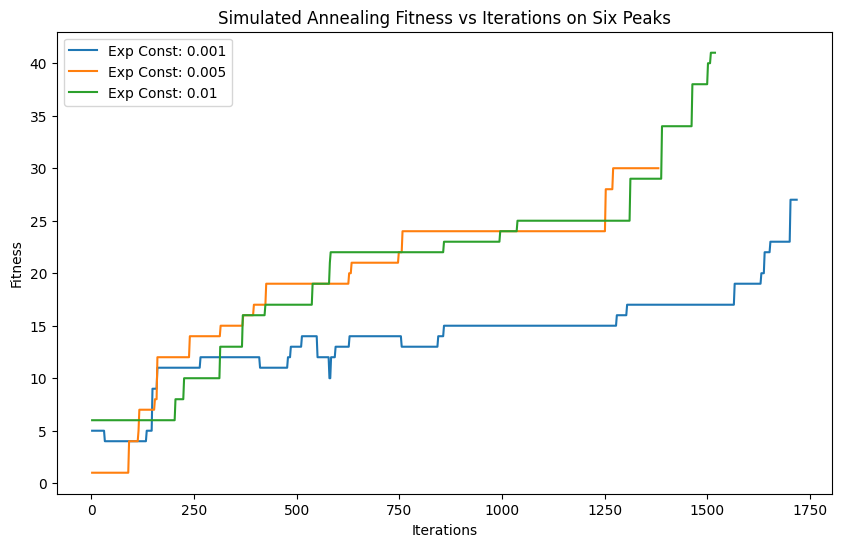}
        \caption{}
        \label{fig:sixpeaksa}
    \end{subfigure}
    \vskip\baselineskip
    \begin{subfigure}[b]{0.49\textwidth}
        \centering
        \includegraphics[width=\textwidth]{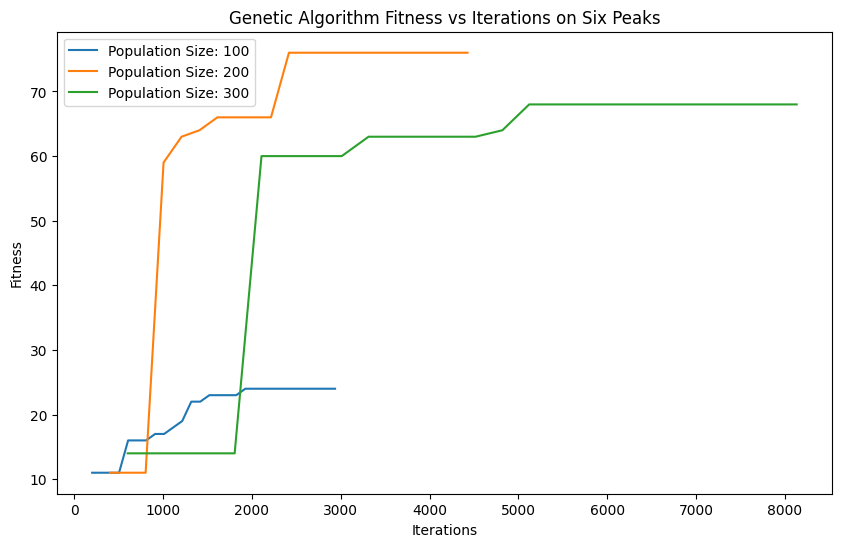}
        \caption{}
        \label{fig:sixpeakga}
    \end{subfigure}
    \hfill
    \begin{subfigure}[b]{0.49\textwidth}
        \centering
        \includegraphics[width=\textwidth]{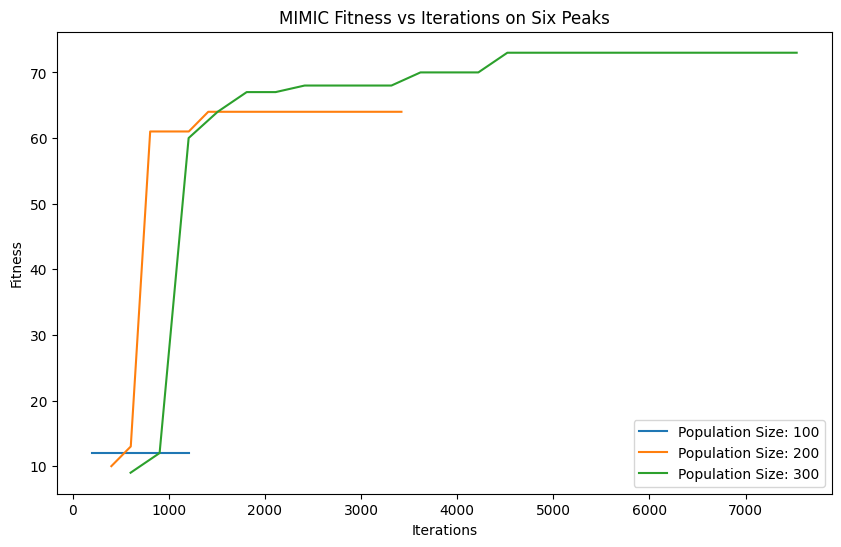}
        \caption{}
        \label{fig:sixpeakmimic}
    \end{subfigure}
    \caption{Fitness vs. iterations for the SixPeaks using different randomized optimization algorithms: (a) RHC, (b) SA, (c) GA, and (d) MIMIC. The plots show the impact of varying key parameters—restarts for RHC, exponential cooling for SA, population sizes for GA, and population sizes for MIMIC—on the fitness progression over iterations.}
    \label{fig:sixpeakgrid}
\end{figure*}

In the SixPeaks problem, where the task is to optimize both peaks of 0’s and 1’s while rewarding strings that achieve long runs, we observe varying performance across algorithms, particularly highlighting the challenges of balancing exploration and exploitation in this more rugged search space. GA and MIMIC again show strong results, with GA reaching an average fitness of 89.2 with a population size of 200 and MIMIC achieving 75.0 with a population size of 300 (Table~\ref{tab:sixpeaks_results}). GA’s ability to reach high fitness efficiently, seen in Figure~\ref{fig:sixpeakga}, reflects its success in quickly navigating this problem’s dual-peak structure, where the balance between exploration and exploitation is critical. MIMIC also demonstrates strong performance, as shown in Figure ~\ref{fig:sixpeakmimic}, converging steadily with different population sizes, though at a slightly slower pace than GA. Meanwhile, RHC struggles even more in the SixPeaks problem compared to previous benchmarks, with fitness values only reaching an average of 9.4 with 10 restarts, highlighting its severe limitations in overcoming the local optima present in the landscape (Table~\ref{tab:sixpeaks_results}, Figure~\ref{fig:sixpeakga}). Compared to the FourPeaks problem, SA’s performance drops marginally in SixPeaks due to the higher level of exploration required.
\begin{table}[h!]
    \centering
    \caption{SixPeaks Problem Results for RHC, SA, GA, and MIMIC with Different Hyperparameters}
    \begin{tabular}{|c|c|c|c|}
        \hline
        \textbf{Algorithm} & \textbf{Param} & \textbf{Average Fitness} & \textbf{Feval} \\ \hline
        
        \textbf{RHC} & Restarts = 0 & 3.2 & 10 \\ \hline
        \textbf{RHC} & Restarts = 5 & 6.6 & 21 \\ \hline
        \textbf{RHC} & Restarts = 10 & 9.4 & 17 \\ \hline
        
        \textbf{SA} & ExpConst = 0.001 & 28.0 & 1000 \\ \hline
        \textbf{SA} & ExpConst = 0.005 & 31.6 & 869 \\ \hline
        \textbf{SA} & ExpConst = 0.01 & 37.0 & 966 \\ \hline
        
        \textbf{GA} & PopSize = 100 & 40.0 & 28 \\ \hline
        \textbf{GA} & PopSize = 200 & 89.2 & 21 \\ \hline
        \textbf{GA} & PopSize = 300 & 73.8 & 26 \\ \hline
        
        \textbf{MIMIC} & PopSize = 100 & 26.4 & 11 \\ \hline
        \textbf{MIMIC} & PopSize = 200 & 73.4 & 16 \\ \hline
        \textbf{MIMIC} & PopSize = 300 & 75.0 & 24 \\ \hline
        
    \end{tabular}
    \label{tab:sixpeaks_results}
\end{table}

\begin{figure*}[htbp]
    \centering
    \begin{subfigure}[b]{0.49\textwidth}
        \centering
        \includegraphics[width=\textwidth]{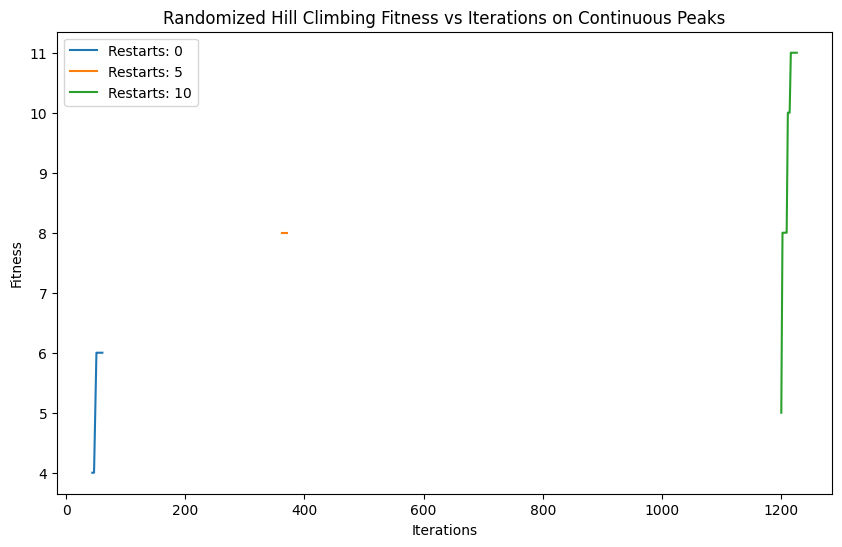}
        \caption{}
        \label{fig:continuouspeakrhc}
    \end{subfigure}
    \hfill
    \begin{subfigure}[b]{0.49\textwidth}
        \centering
        \includegraphics[width=\textwidth]{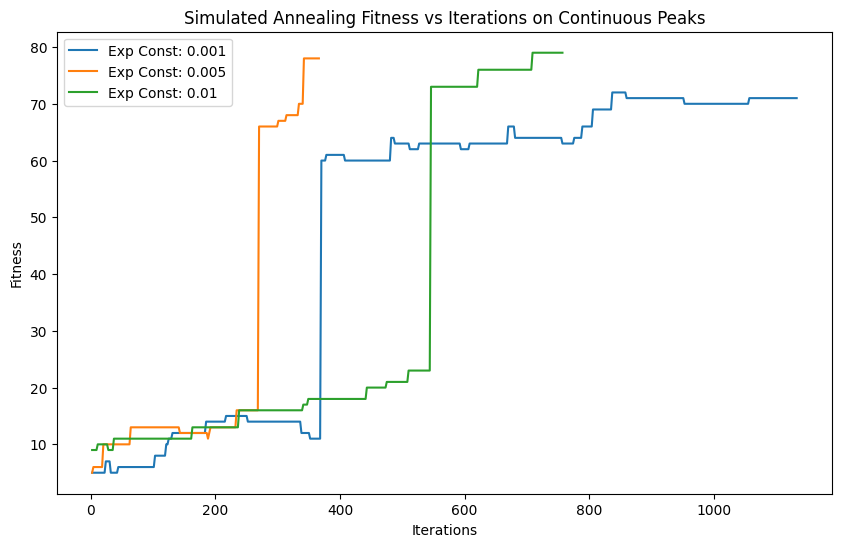}
        \caption{}
        \label{fig:continuouspeaksa}
    \end{subfigure}
    \vskip\baselineskip
    \begin{subfigure}[b]{0.49\textwidth}
        \centering
        \includegraphics[width=\textwidth]{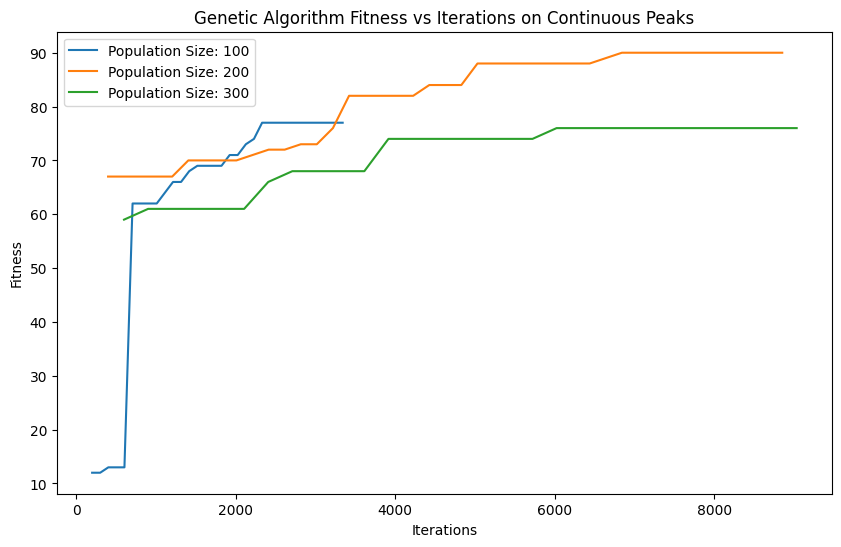}
        \caption{}
        \label{fig:continuouspeakga}
    \end{subfigure}
    \hfill
    \begin{subfigure}[b]{0.49\textwidth}
        \centering
        \includegraphics[width=\textwidth]{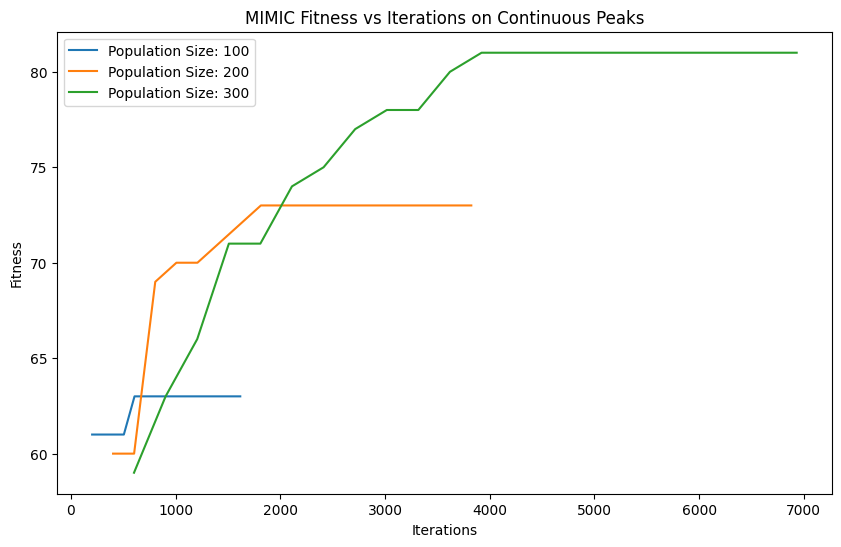}
        \caption{}
        \label{fig:continuouspeakmimic}
    \end{subfigure}
    \caption{Fitness vs. iterations for the ContinuousPeaks using different randomized optimization algorithms: (a) RHC, (b) SA, (c) GA, and (d) MIMIC. The plots show the impact of varying key parameters—restarts for RHC, exponential cooling for SA, population sizes for GA, and population sizes for MIMIC—on the fitness progression over iterations.}
    \label{fig:continuouspeakgrid}
\end{figure*}

In the ContinuousPeaks problem, where the goal is to balance between long sequences of 0’s and 1’s, GA and MIMIC continue to excel, with GA achieving an average fitness of 82.4 for a population size of 200 and MIMIC reaching 80.8 with a population size of 300 (Table~\ref{tab:continuouspeaks_results}). GA’s efficiency in this problem, as seen in Figure~\ref{fig:continuouspeakga}, is due to its ability to exploit the structured peaks in the search space, allowing for rapid convergence to high-quality solutions. MIMIC also performs well, steadily improving with larger population sizes (Figure~\ref{fig:continuouspeakmimic}), although it converges slightly more slowly compared to GA. RHC shows poor performance once again, with an average fitness of only 23.8 even with 10 restarts, underscoring its limitations in escaping local optima in this landscape (Table~\ref{tab:continuouspeaks_results}, Figure~\ref{fig:continuouspeakrhc}). Interestingly, SA performs much better in this problem compared to previous benchmarks, achieving an average fitness of 77.2 with an exponential constant of 0.001, likely due to the problem’s more defined peaks, which align well with SA's probabilistic exploration mechanism (Table~\ref{tab:continuouspeaks_results}, Figure~\ref{fig:continuouspeaksa}) or possibly due to a sheer stroke of luck.
\begin{table}[h!]
    \centering
    \caption{ContinuousPeaks Problem Results for RHC, SA, GA, and MIMIC with Different Hyperparameters}
    \begin{tabular}{|c|c|c|c|}
        \hline
        \textbf{Algorithm} & \textbf{Param} & \textbf{Average Fitness} & \textbf{Feval} \\ \hline
        
        \textbf{RHC} & Restarts = 0 & 8.6 & 16 \\ \hline
        \textbf{RHC} & Restarts = 5 & 9.6 & 10 \\ \hline
        \textbf{RHC} & Restarts = 10 & 23.8 & 24 \\ \hline
        
        \textbf{SA} & ExpConst = 0.001 & 77.2 & 712 \\ \hline
        \textbf{SA} & ExpConst = 0.005 & 64.6 & 222 \\ \hline
        \textbf{SA} & ExpConst = 0.01 & 71.8 & 483 \\ \hline
        
        \textbf{GA} & PopSize = 100 & 71.0 & 32 \\ \hline
        \textbf{GA} & PopSize = 200 & 82.4 & 43 \\ \hline
        \textbf{GA} & PopSize = 300 & 81.6 & 29 \\ \hline
        
        \textbf{MIMIC} & PopSize = 100 & 68.2 & 15 \\ \hline
        \textbf{MIMIC} & PopSize = 200 & 74.4 & 18 \\ \hline
        \textbf{MIMIC} & PopSize = 300 & 80.8 & 22 \\ \hline
        
    \end{tabular}
    \label{tab:continuouspeaks_results}
\end{table}

\subsection{Permutation Problems}

\begin{figure*}[htbp]
    \centering
    \begin{subfigure}[b]{0.49\textwidth}
        \centering
        \includegraphics[width=\textwidth]{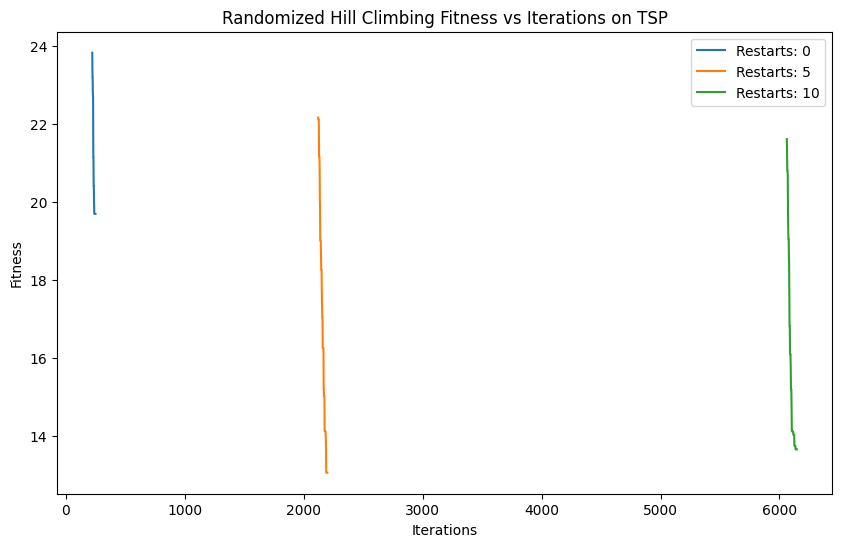}
        \caption{}
        \label{fig:tsprhc}
    \end{subfigure}
    \hfill
    \begin{subfigure}[b]{0.49\textwidth}
        \centering
        \includegraphics[width=\textwidth]{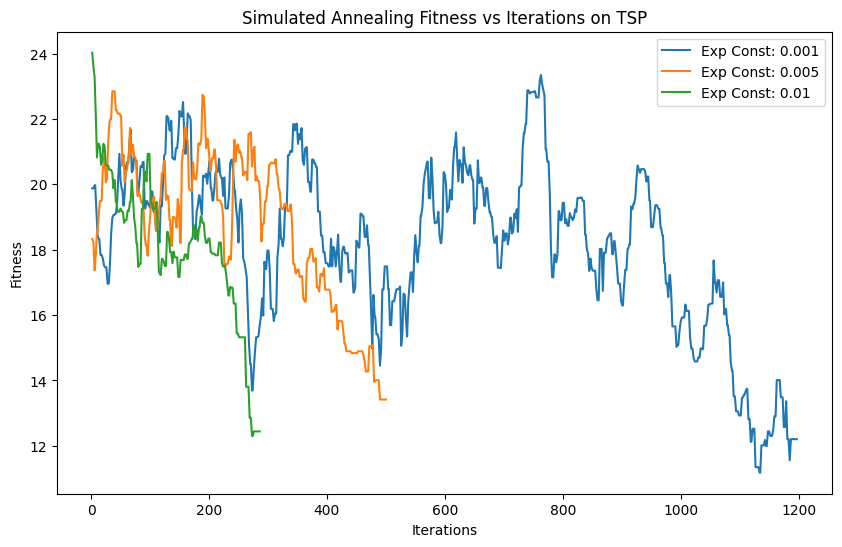}
        \caption{}
        \label{fig:tspsa}
    \end{subfigure}
    \vskip\baselineskip
    \begin{subfigure}[b]{0.49\textwidth}
        \centering
        \includegraphics[width=\textwidth]{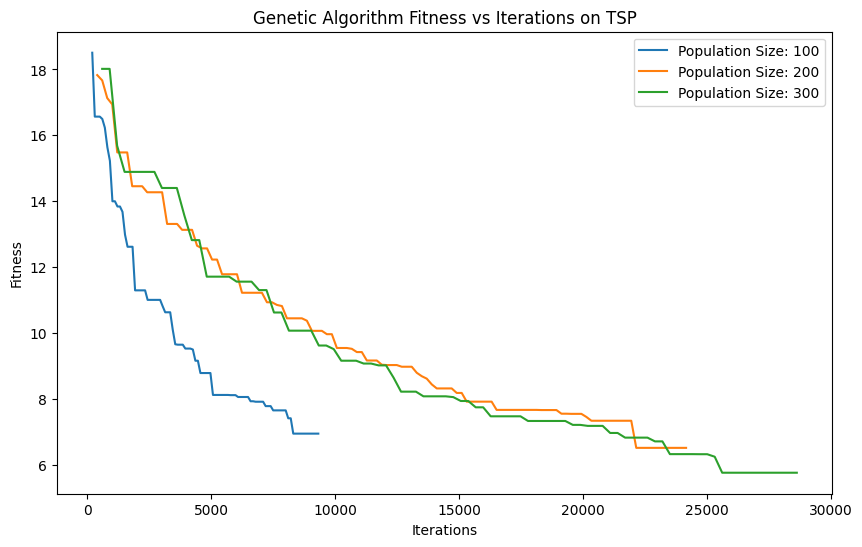}
        \caption{}
        \label{fig:tspga}
    \end{subfigure}
    \hfill
    \begin{subfigure}[b]{0.49\textwidth}
        \centering
        \includegraphics[width=\textwidth]{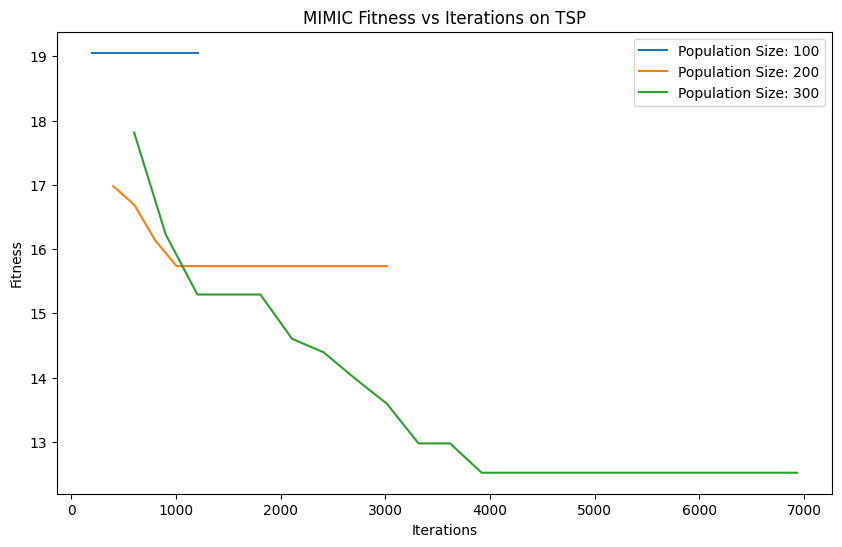}
        \caption{}
        \label{fig:tspmimic}
    \end{subfigure}
    \caption{Fitness vs. iterations for the Traveling Salesman Problem using different randomized optimization algorithms: (a) RHC, (b) SA, (c) GA, and (d) MIMIC. The plots show the impact of varying key parameters—restarts for RHC, exponential cooling for SA, population sizes for GA, and population sizes for MIMIC—on the fitness progression over iterations.}
    \label{fig:tspgrid}
\end{figure*}

In the Traveling Salesman Problem (TSP), where the goal is to find the shortest possible route through a set of cities, the performance of the algorithms reflects the challenge of navigating a complex, highly constrained search space. RHC performs relatively well, ony behind MIMIC, achieving an average fitness of 16.05 with 0 restarts and 13.75 with 5 restarts (Table~\ref{tab:tsp_results}). This suggests that RHC can escape local optima to some extent with no restarts, but additional restarts do not significantly improve the result. However, with 10 restarts, the average fitness drops to 11.52, showing that excessive restarts can hurt performance in this scenario. GA demonstrates a more consistent convergence pattern than both RHC and SA. With a population size of 100, it achieves an average fitness of 8.80, and with a population size of 200, it improves to 9.30. When the population size is increased to 300, the average fitness is further improved to 9.62. MIMIC outperforms all other algorithms (Figure~\ref{fig:tspgrid}), achieving the highest average fitness of 18.47 with a population size of 100. With a population size of 200, MIMIC achieves a slightly lower average fitness of 17.24, and with a population size of 300, it achieves 16.75. This indicates that MIMIC is highly effective at exploring promising regions of the solution space using its probabilistic model, but larger population sizes can lead to diminishing returns. MIMIC converges faster than GA, particularly with smaller population sizes, and efficiently solves the TSP.
\begin{table}[h!]
    \centering
    \caption{Traveling Salesman Problem Results for RHC, SA, GA, and MIMIC with Different Hyperparameters}
    \begin{tabular}{|c|c|c|c|}
        \hline
        \textbf{Algorithm} & \textbf{Param} & \textbf{Average Fitness} & \textbf{Feval} \\ \hline
        
        \textbf{RHC} & Restarts = 0 & 16.05 & 21 \\ \hline
        \textbf{RHC} & Restarts = 5 & 13.78 & 63 \\ \hline
        \textbf{RHC} & Restarts = 10 & 11.52 & 66 \\ \hline
        
        \textbf{SA} & ExpConst = 0.001 & 13.91 & 761 \\ \hline
        \textbf{SA} & ExpConst = 0.005 & 13.40 & 333 \\ \hline
        \textbf{SA} & ExpConst = 0.01 & 13.52 & 191 \\ \hline
        
        \textbf{GA} & PopSize = 100 & 8.80 & 91 \\ \hline
        \textbf{GA} & PopSize = 200 & 6.91 & 119 \\ \hline
        \textbf{GA} & PopSize = 300 & 6.29 & 94 \\ \hline
        
        \textbf{MIMIC} & PopSize = 100 & 18.47 & 11 \\ \hline
        \textbf{MIMIC} & PopSize = 200 & 15.90 & 14 \\ \hline
        \textbf{MIMIC} & PopSize = 300 & 13.31 & 22 \\ \hline
        
    \end{tabular}
    \label{tab:tsp_results}
\end{table}

\begin{figure*}[htbp]
    \centering
    \begin{subfigure}[b]{0.49\textwidth}
        \centering
        \includegraphics[width=\textwidth]{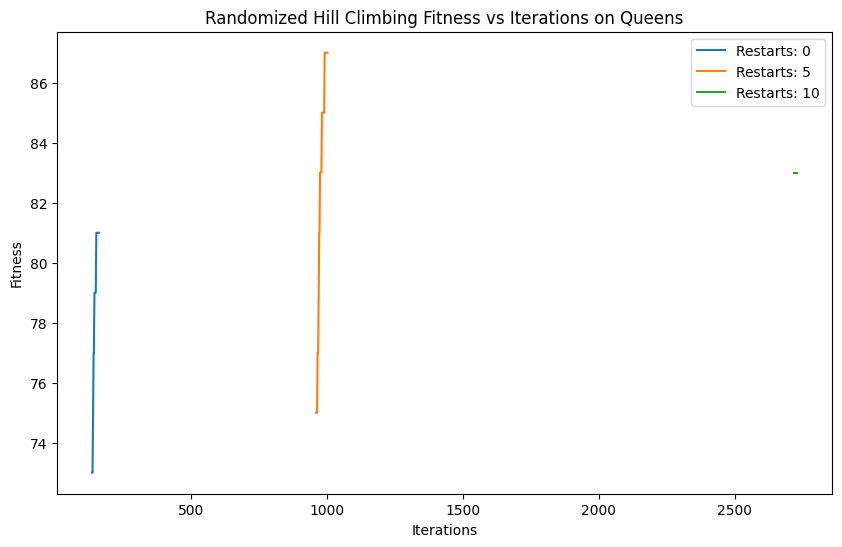}
        \caption{}
        \label{fig:queensrhc}
    \end{subfigure}
    \hfill
    \begin{subfigure}[b]{0.49\textwidth}
        \centering
        \includegraphics[width=\textwidth]{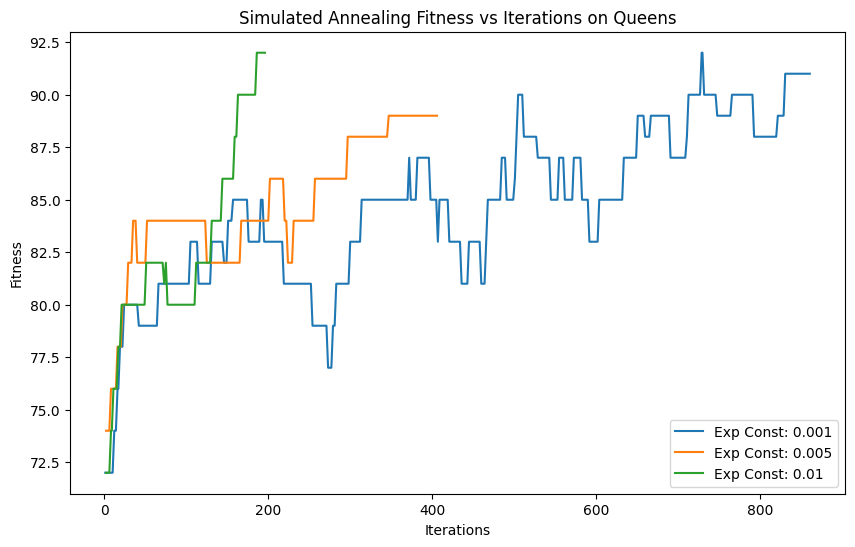}
        \caption{}
        \label{fig:queenssa}
    \end{subfigure}
    \vskip\baselineskip
    \begin{subfigure}[b]{0.49\textwidth}
        \centering
        \includegraphics[width=\textwidth]{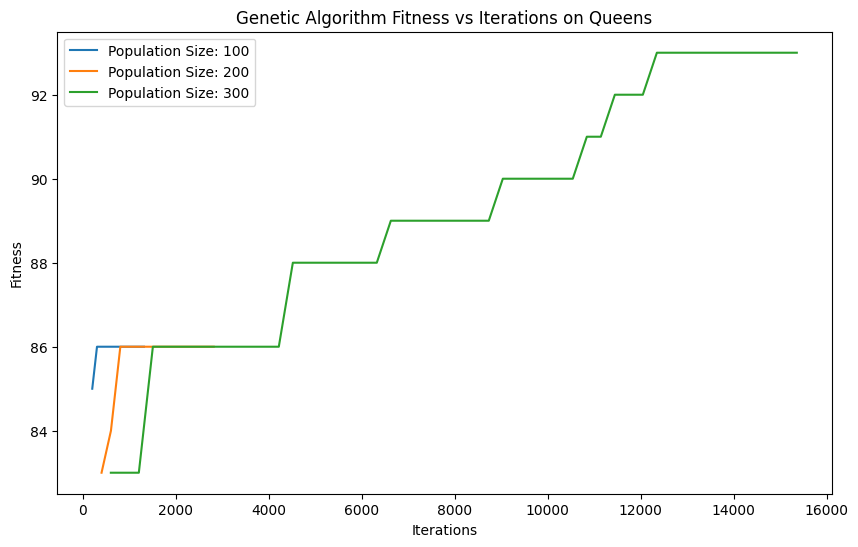}
        \caption{}
        \label{fig:queensga}
    \end{subfigure}
    \hfill
    \begin{subfigure}[b]{0.49\textwidth}
        \centering
        \includegraphics[width=\textwidth]{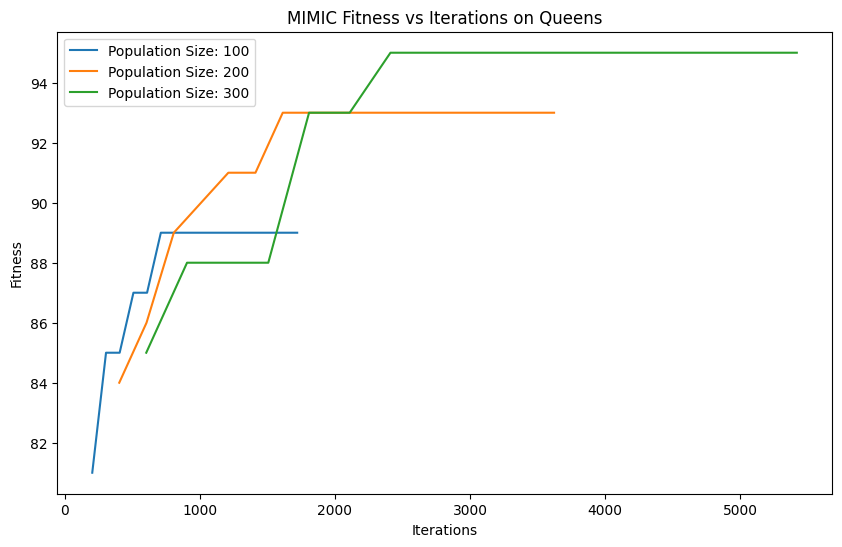}
        \caption{}
        \label{fig:queensmimic}
    \end{subfigure}
    \caption{Fitness vs. iterations for the Queens Problem using different randomized optimization algorithms: (a) RHC, (b) SA, (c) GA, and (d) MIMIC. The plots show the impact of varying key parameters—restarts for RHC, exponential cooling for SA, population sizes for GA, and population sizes for MIMIC—on the fitness progression over iterations.}
    \label{fig:queensgrid}
\end{figure*}

In the Queens problem, where the objective is to place queens on a chessboard such that no two queens threaten each other, the results show that all algorithms perform significantly better than in previous permutation problems, reflecting the structured nature of the solution space. MIMIC performs best, with an average fitness of 95.4 for a population size of 300, quickly converging to near-optimal solutions (Table~\ref{tab:queens_results}, Figure~\ref{fig:queensmimic}). This highlights MIMIC’s strong ability to capture and exploit dependencies between variables in the Queens problem. GA also perform well, achieving an average fitness of 90.1 with a population size of 100, though it takes longer to converge compared to MIMIC (Figure~\ref{fig:queensga}). SA shows improved performance here compared to the TSP, with an average fitness of 90.6 at an exponential constant of 0.01, although its convergence is less stable, as shown in Figure~\ref{fig:queenssa}. The probabilistic acceptance of worse solutions helps SA escape local optima, but it introduces variability, making it less reliable than GA and MIMIC in finding the optimal solution. RHC performs reasonably well for a local search algorithm, reaching an average fitness of 85.6 with 5 restarts, but it still lags behind the other methods, as it struggles to maintain global exploration. 
\begin{table}[h!]
    \centering
    \caption{Queens Problem Results for RHC, SA, GA, and MIMIC with Different Hyperparameters}
    \begin{tabular}{|c|c|c|c|}
        \hline
        \textbf{Algorithm} & \textbf{Param} & \textbf{Average Fitness} & \textbf{Feval} \\ \hline
        
        \textbf{RHC} & Restarts = 0 & 81.4 & 22 \\ \hline
        \textbf{RHC} & Restarts = 5 & 85.6 & 37 \\ \hline
        \textbf{RHC} & Restarts = 10 & 85.0 & 10 \\ \hline
        
        \textbf{SA} & ExpConst = 0.001 & 89.8 & 587 \\ \hline
        \textbf{SA} & ExpConst = 0.005 & 89.2 & 277 \\ \hline
        \textbf{SA} & ExpConst = 0.01 & 90.6 & 130 \\ \hline
        
        \textbf{GA} & PopSize = 100 & 91.0 & 12 \\ \hline
        \textbf{GA} & PopSize = 200 & 86.8 & 13 \\ \hline
        \textbf{GA} & PopSize = 300 & 91.0 & 50 \\ \hline
        
        \textbf{MIMIC} & PopSize = 100 & 91.4 & 16 \\ \hline
        \textbf{MIMIC} & PopSize = 200 & 94.2 & 17 \\ \hline
        \textbf{MIMIC} & PopSize = 300 & 95.4 & 17 \\ \hline
        
    \end{tabular}
    \label{tab:queens_results}
\end{table}

\subsection{Combinatorial Problems}
\begin{figure*}[htbp]
    \centering
    \begin{subfigure}[b]{0.49\textwidth}
        \centering
        \includegraphics[width=\textwidth]{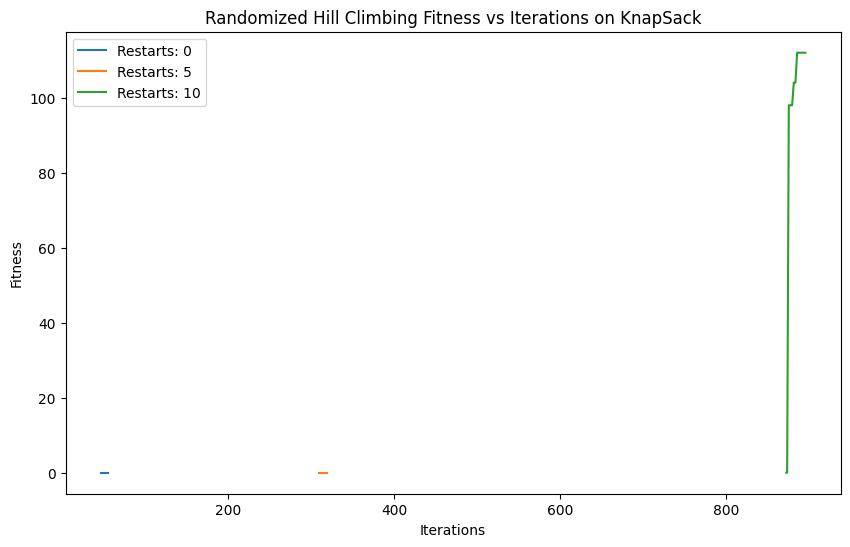}
        \caption{}
        \label{fig:knapsackrhc}
    \end{subfigure}
    \hfill
    \begin{subfigure}[b]{0.49\textwidth}
        \centering
        \includegraphics[width=\textwidth]{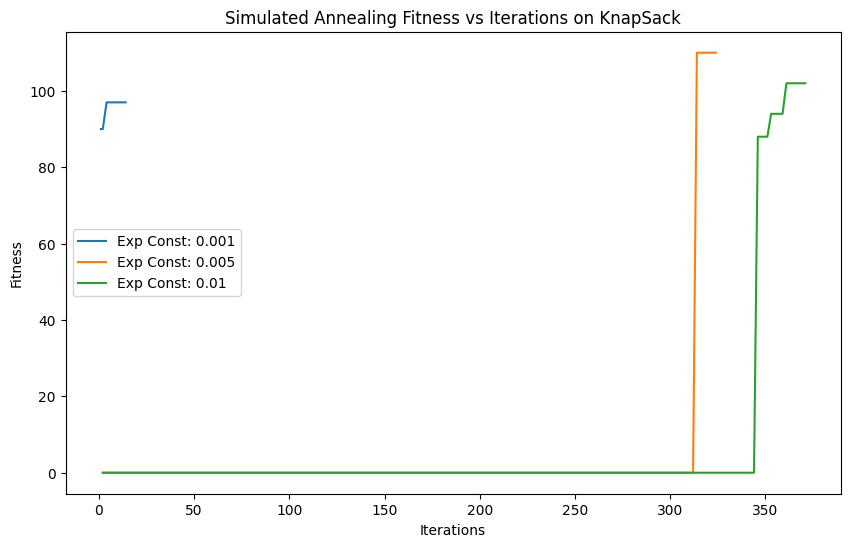}
        \caption{}
        \label{fig:knapsacksa}
    \end{subfigure}
    \vskip\baselineskip
    \begin{subfigure}[b]{0.49\textwidth}
        \centering
        \includegraphics[width=\textwidth]{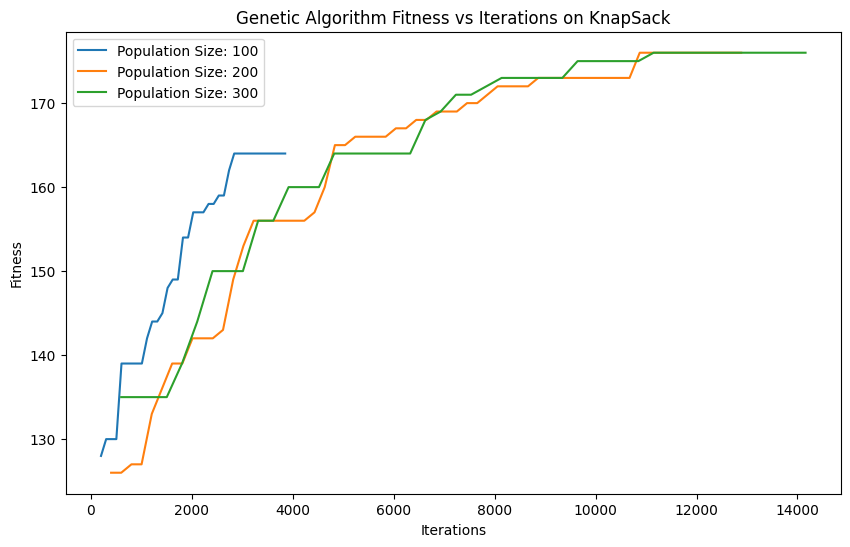}
        \caption{}
        \label{fig:knapsackga}
    \end{subfigure}
    \hfill
    \begin{subfigure}[b]{0.49\textwidth}
        \centering
        \includegraphics[width=\textwidth]{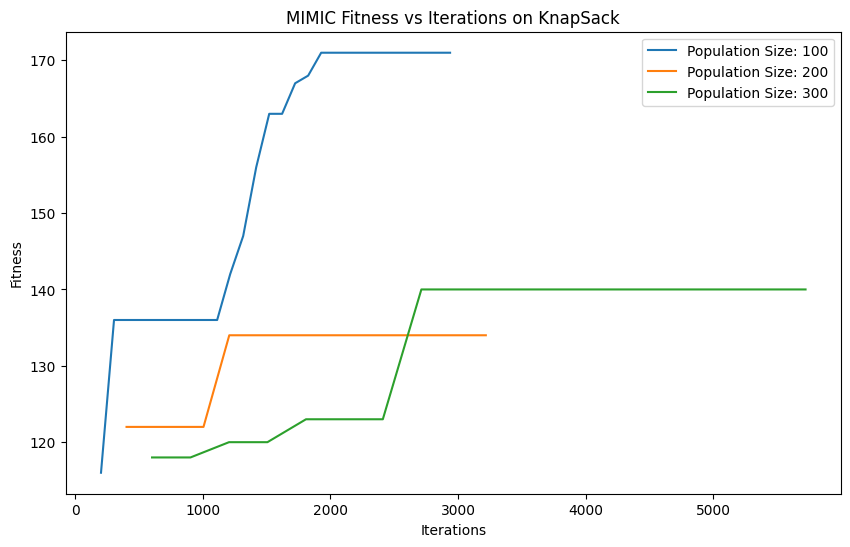}
        \caption{}
        \label{fig:knapsackmimic}
    \end{subfigure}
    \caption{Fitness vs. iterations for the Knapsack Problem using different randomized optimization algorithms: (a) RHC, (b) SA, (c) GA, and (d) MIMIC. The plots show the impact of varying key parameters—restarts for RHC, exponential cooling for SA, population sizes for GA, and population sizes for MIMIC—on the fitness progression over iterations.}
    \label{fig:knapsackgrid}
\end{figure*}

In the Knapsack problem, where the goal is to maximize the value of selected items while staying within a weight limit, the results demonstrate how each algorithm handles the combinatorial complexity. GA perform the best, with an average fitness of 175.8 for a population size of 300 (Table ~\ref{tab:knapsack_results}). GA’s crossover and mutation operations help it efficiently explore the solution space, as seen in Figure~\ref{fig:knapsackga}, where the fitness improves steadily, demonstrating its capacity to balance exploration and exploitation effectively. MIMIC also shows strong performance, particularly with a population size of 100, achieving an average fitness of 162.0. However, as shown in Figure~\ref{fig:knapsackmimic}, MIMIC's fitness progression tends to plateau after an initial steep climb, especially for larger population sizes, indicating that the algorithm struggles to continue improving beyond a certain point in this combinatorial problem. SA performs moderately well, reaching an average fitness of 111.0 with an exponential constant of 0.001. However, its performance fluctuates, as seen in Figure~\ref{fig:knapsacksa}, where SA either converges rapidly or lags behind, depending on the cooling schedule. RHC performs poorly compared to the other methods, with an average fitness of only 91.2 with 10 restarts. RHC’s inability to effectively escape local optima and explore larger portions of the solution space hampers its performance in the more complex search space of the Knapsack problem, a limitation that has been consistent across the other problems analyzed.
\begin{table}[h!]
    \centering
    \caption{Knapsack Problem Results for RHC, SA, GA, and MIMIC with Different Hyperparameters}
    \begin{tabular}{|c|c|c|c|}
        \hline
        \textbf{Algorithm} & \textbf{Param} & \textbf{Average Fitness} & \textbf{Feval} \\ \hline
        
        \textbf{RHC} & Restarts = 0 & 19.2 & 10 \\ \hline
        \textbf{RHC} & Restarts = 5 & 46.0 & 10 \\ \hline
        \textbf{RHC} & Restarts = 10 & 91.2 & 21 \\ \hline
        
        \textbf{SA} & ExpConst = 0.001 & 111.0 & 13 \\ \hline
        \textbf{SA} & ExpConst = 0.005 & 114.6 & 167 \\ \hline
        \textbf{SA} & ExpConst = 0.01 & 84.8 & 196 \\ \hline
        
        \textbf{GA} & PopSize = 100 & 167.4 & 37 \\ \hline
        \textbf{GA} & PopSize = 200 & 172.2 & 63 \\ \hline
        \textbf{GA} & PopSize = 300 & 175.8 & 46 \\ \hline
        
        \textbf{MIMIC} & PopSize = 100 & 145.0 & 28 \\ \hline
        \textbf{MIMIC} & PopSize = 200 & 129.8 & 15 \\ \hline
        \textbf{MIMIC} & PopSize = 300 & 136.8 & 18 \\ \hline
        
    \end{tabular}
    \label{tab:knapsack_results}
\end{table}

\subsection{Cross-Group Performance Analysis of Optimization Algorithms}
\begin{figure}[h!]
    \centering
    \includegraphics[width=0.46\textwidth]{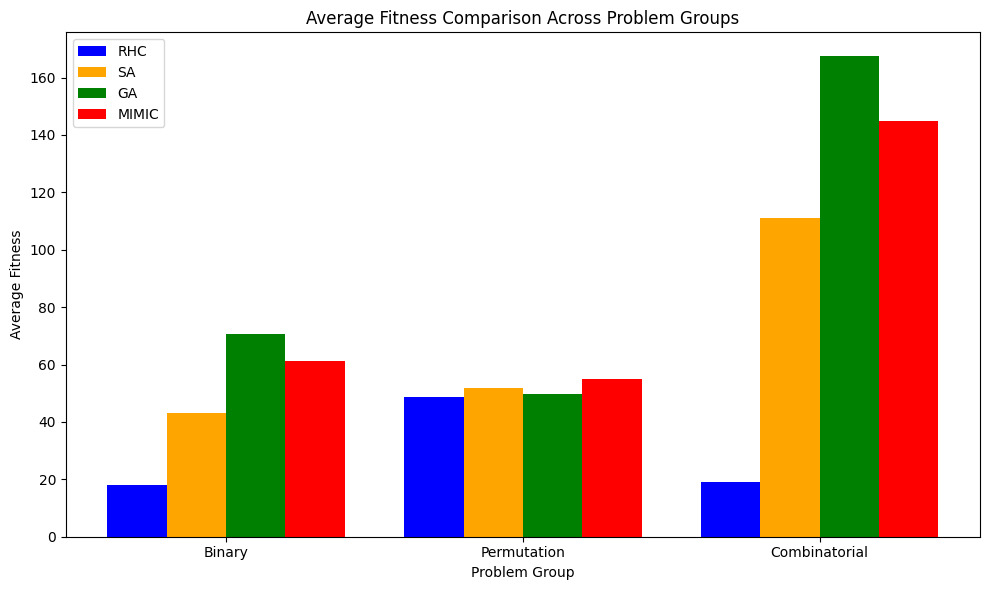}
    \caption{Average Fitness Comparison Across Problem Groups (Binary, Permutation, Combinatorial) for RHC, SA, GA, and MIMIC}
    \label{fig:average_fitness_comparison}
\end{figure}

In this section, we compare the performance of RHC, SA, GA, and MIMIC across binary, permutation, and combinatorial problem groups, highlighting the distinct trends in efficiency, computational cost, and solution quality that emerge. As seen in Figure \ref{fig:average_fitness_comparison}, MIMIC and GA perform exceptionally well across all problem groups, particularly excelling in the combinatorial group, where MIMIC achieved over 80\% of the total possible fitness for the Knapsack problem, and GA reached 86\%. For the permutation group, MIMIC showed a similar trend with fitness values hovering around 90\% in problems like Queens, while GA exhibited consistent, though slightly slower, performance. The binary group had a simpler fitness landscape, allowing both MIMIC and GA to converge efficiently, although the binary problems generally presented fewer local optima and required less sophisticated search mechanisms than the combinatorial and permutation problems.

\begin{figure}[h!]
    \centering
    \includegraphics[width=0.46\textwidth]{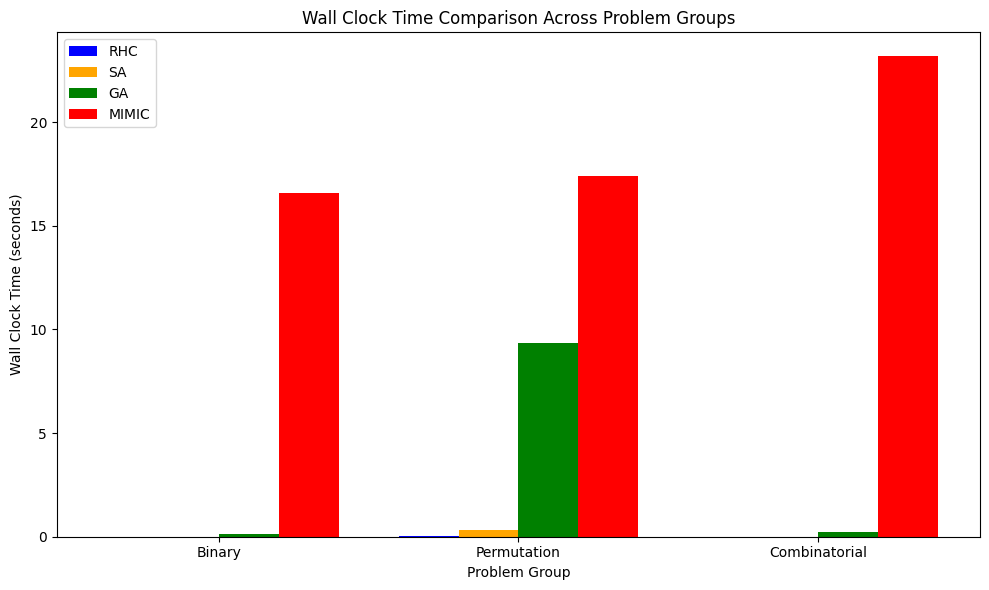}
    \caption{Wall Clock Time Comparison Across Problem Groups (Binary, Permutation, Combinatorial) for RHC, SA, GA, and MIMIC}
    \label{fig:wall_clock_time_comparison}
\end{figure}

However, a deeper comparison reveals trade-offs between solution quality and computational costs. As shown in Figure \ref{fig:wall_clock_time_comparison}, MIMIC incurs higher computational cost in terms of wall clock time across all groups, particularly in the permutation and combinatorial problems. For example, in the binary problem group, MIMIC's higher wall clock time was more pronounced compared to GA, which maintained moderate time efficiency while still producing competitive fitness results. This highlights MIMIC’s strength in consistently achieving high solution quality, albeit at the cost of increased computation, making it more suitable for cases where accuracy is paramount, and computational resources are less constrained. On the other hand, GA demonstrated a more balanced approach across groups, achieving high-quality solutions with significantly lower computational overhead than MIMIC, making it more appropriate for situations where a compromise between solution quality and time efficiency is needed.

\begin{figure}[h!]
    \centering
    \includegraphics[width=0.46\textwidth]{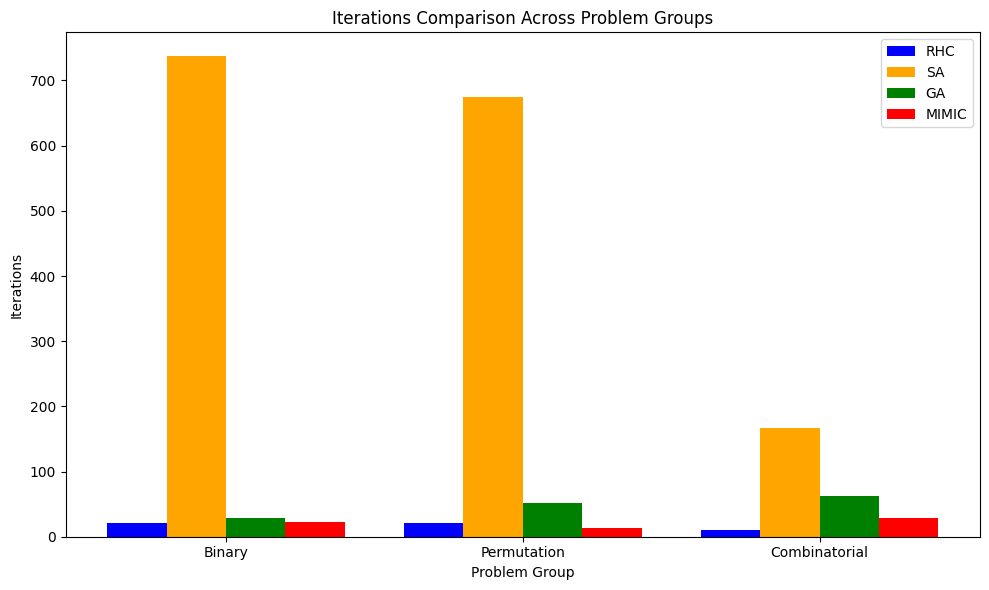}
    \caption{Function evaluation Comparison Across Problem Groups (Binary, Permutation, Combinatorial) for RHC, SA, GA, and MIMIC}
    \label{fig:iterations_comparison}
\end{figure}

RHC and SA, as expected, struggled to maintain competitive performance across the groups. As shown in Figure \ref{fig:iterations_comparison}, SA required the highest number of feval, particularly in binary and permutation problems, reflecting its difficulty in efficiently exploring simpler landscapes where local optima are less challenging to escape. RHC consistently underperformed, achieving less than 60\% of the total possible fitness in all groups, with its performance particularly poor in the combinatorial problems. Its inability to explore the search space effectively and its susceptibility to getting trapped in local optima, as shown by the relatively few feval in Figure \ref{fig:iterations_comparison}, make it the least effective algorithm across all problem types. We summarise the results of our experiments in Table~\ref{tab:resultsummary}.

\begin{table*}[h!]
\centering
\caption{Algorithm selection based on problem group, fitness achieved, and computational cost.}
\begin{tabular}{|p{2cm}|p{1cm}|p{2cm}|p{2cm}|p{4cm}|p{4cm}|}
\hline
\textbf{Problem Group}       & \textbf{Best Algorithm} & \textbf{Fitness Achieved (\% of Optimum)} & \textbf{Computational Cost (Time/Iterations)} & \textbf{When to Use}                                                                                       & \textbf{Trade-offs}                                                                                               \\ \hline
\textbf{Binary}              & GA, MIMIC               & 90-95\%                                    & Moderate (GA), High (MIMIC)                    & Use GA for most binary problems due to its lower time cost, MIMIC when accuracy is critical                  & MIMIC gives slightly better fitness, but GA is faster overall                                                      \\ \hline
\textbf{Permutation}         & MIMIC, GA               & 85-90\%                                    & High (MIMIC), Moderate (GA)                    & Use MIMIC for complex problems like Queens where interdependencies are important, GA for faster convergence  & MIMIC provides better solutions but at significantly higher computational cost                                     \\ \hline
\textbf{Combinatorial}       & GA                      & 85-90\%                                    & Moderate                                       & Use GA for large combinatorial problems with complex constraints                                             & MIMIC can be used in place of GA but at significant computation cost with not much difference in solution quality             \\ \hline
\textbf{Low Computational Cost} & RHC                   & \textless{}60\%                            & Low                                            & Use RHC only when computation resources are limited and solution quality is not critical                     & Very poor solution quality makes RHC unsuitable for most large or complex problems                                 \\ \hline
\textbf{High Accuracy Need}  & MIMIC                   & 85-95\%                                    & High                                           & Use MIMIC when solution quality is the highest priority, regardless of computational cost                    & High computational cost, particularly in larger and more complex problem types                                     \\ \hline
\end{tabular}
\label{tab:resultsummary}
\end{table*}
\section{Conclusion}
In this study, we assessed the performance of four optimization algorithms—RHC, SA, GA, and MIMIC—by analyzing their underlying mechanisms and comparing their strengths and weaknesses. These algorithms were tested across three distinct problem categories: binary (including OneMax, FlipFlop, FourPeaks, SixPeaks, and ContinuousPeaks), permutation (Traveling Salesman and N-Queens), and combinatorial (Knapsack). The performance and behavior of each algorithm were evaluated through multiple experimental trials. Results showed that while MIMIC and GA demonstrated superior performance for all problem group, they also came with significantly higher computational costs. Ongoing research aims to develop more efficient algorithms that maintain high performance while reducing computational overhead. Additionally, a specific study was conducted to assess the robustness and resilience of each algorithm in applications such as planning and scheduling. Preliminary work exploring the concept of algorithmic resilience in these contexts can be found in earlier research \cite{han2016resilience, raj2014resilience}. Another avenue for future work is to explore how different coding schemes may affect the performance of optimization algorithms. Literature suggests that coding schemes can significantly impact efficiency, as shown in \cite{9348654}. where a proposed coding scheme reduced the number of iterations by 17\%.
\bibliography{references}
\begin{IEEEbiography}[{\includegraphics[width=1in,height=1.25in,clip,keepaspectratio]{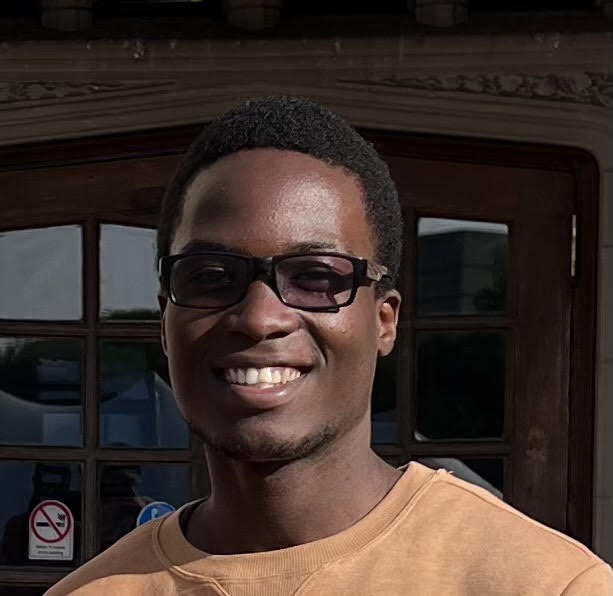}}]{Jethro Odeyemi} is currently a Ph.D. student in Biomedical Engineering at the University of Saskatchewan, Saskatoon, SK, Canada. His research focuses on medical robotics, machine learning, and human-computer interaction. He holds an M.Sc. in Computer Science with a specialization in Machine Learning from Georgia Institute of Technology, and an M.Sc. in Biomedical Engineering from the University of Saskatchewan. He also earned a B.Eng. in Mechatronics Engineering from the Federal University of Agriculture, Abeokuta, Nigeria, and holds an Advanced Level Certificate in Mathematics from the University of Cambridge. He is an Engineer-in-Training with the Association of Professional Engineers and Geoscientists of Saskatchewan (APEGS).
\end{IEEEbiography}

\begin{IEEEbiography}[{\includegraphics[width=1in,height=1.25in,clip,keepaspectratio]{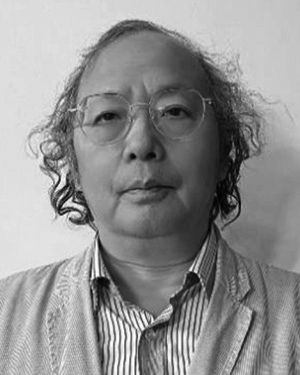}}]{Wenjun Zhang} (Senior Member, IEEE) received the Ph.D. degree from Delft University of Technology, Delft, The Netherlands, in 1994.,He is currently a Full Professor with the Department of Mechanical Engineering and the Division of Biomedical Engineering, University of Saskatchewan, Saskatoon, SK, Canada. He has published more than 300 technical papers in peer-refereed journals and more than 200 technical papers in peer-refereed conference proceedings.,Dr. Zhang is currently a Senior Editor for the IEEE/ASME Transactions on Mechatronics. He is a Fellow of the Canadian Academy of Engineering owing to his outstanding work on resilience engineering.
\end{IEEEbiography}
\end{document}